\documentclass[manuscript]{acmart}
\usepackage[caption=false,font=footnotesize,labelfont=rm,textfont=rm]{subfig}
\AtBeginDocument{%
  }

\setcopyright{acmlicensed}
\copyrightyear{2018}
\acmYear{2018}
\acmDOI{XXXXXXX.XXXXXXX}
\acmConference[Conference acronym 'XX]{Make sure to enter the correct
  conference title from your rights confirmation email}{June 03--05,
  2018}{Woodstock, NY}
\acmISBN{978-1-4503-XXXX-X/2018/06}

\begin{document}

\title{Tokenizing Motion: A Generative Approach for Scene Dynamics Compression}

\author{Shanzhi Yin}
\email{shanzhyin3-c@my.cityu.edu.hk}
\author{Zihan Zhang}
\email{zhzhang38-c@my.cityu.edu.hk}
\affiliation{%
  \institution{City University of Hong Kong}
  \city{Kowloon}
  \state{Hong Kong}
  \country{China}
}

\author{Bolin Chen}
\email{chenbolin.chenboli@alibaba-inc.com}

\affiliation{%
  \institution{Alibaba DAMO Academy and Hupan Lab}
  \city{Hangzhou}
  \country{China}}

\author{Ru-Ling Liao}
\email{ruling.lrl@alibaba-inc.com}
\affiliation{%
  \institution{Alibaba DAMO Academy}
  \city{Sunnyvale}
  \country{USA}}

\author{Shiqi Wang}
\email{shiqwang@my.cityu.edu.hk}
\affiliation{%
  \institution{City University of Hong Kong}
  \city{Kowloon}
  \state{Hong Kong}
  \country{China}
}

\author{Yan Ye}
\email{yan.ye@alibaba-inc.com}
\affiliation{%
  \institution{Alibaba DAMO Academy}
  \city{Sunnyvale}
  \country{USA}}

\renewcommand{\shortauthors}{Yin et al.}

\begin{abstract}
  This paper proposes a novel generative video compression framework that leverages motion pattern priors, derived from subtle dynamics in common scenes (e.g., swaying flowers or a boat drifting on water), rather than relying on video content priors (e.g., talking faces or human bodies). These compact motion priors enable a new approach to ultra-low bit-rate communication while achieving high-quality reconstruction across diverse scene contents.
At the encoder side, motion priors can be streamlined into compact representations via a dense-to-sparse transformation. At the decoder side, these priors facilitate the reconstruction of scene dynamics using an advanced flow-driven diffusion model.
Experimental results illustrate that the proposed method can achieve superior rate-distortion performance and outperform the state-of-the-art conventional video codec Enhanced Compression Model~(ECM) on scene dynamics sequences. The project page can be found at \url{https://github.com/xyzysz/GNVDC}.
\end{abstract}




\maketitle

\section{Introduction}
Motion is the eternal melody of the world. Efficiently characterizing motion patterns and reducing temporal redundancy is of vital importance to video coding techniques. With the recent advancements of ``Artificial Intelligence Generated Content~(AIGC)'' techniques, generative video coding~(GVC) is proposed to reduce the transmission redundancy and achieve ultra-low bit-rate coding by leveraging the statistical regularities in particular video contents such as human faces~\cite{gfvc-review}. Specifically, the motion flow can be predicted from compact feature representations that are extracted from the key frame and inter frames. Then, the strong generation ability of the deep generative model can guarantee visual-pleasing reconstruction by animating the key frame with the corresponding optical flows~\cite{generativevisualcompressionreview}. Such a design can warrant superior Rate-Distortion~(RD) performance and outperform the latest conventional video coding standard Versatile Video Coding~(VVC) with a large margin in terms of perceptual metrics~\cite{cttr}.

Existing GVC methods usually use explicit representations with physical meaning as the precursors of motion representations. For example, Deep Animation Codec~\cite{dac} adopts 2D key-points representations~\cite{fomm} to build generative video-conferencing codec, Face Video-to-Video Synthesis~\cite{fv2v} further leverages 3D key-points for free-view talking-head synthesis, and an extreme generative human-oriented video codec~\cite{mraa-codec} is built by compressing the articulated motion representations~\cite{mraa}. Meanwhile, implicit features are also explored for direct motion representations. Compact Feature Temporal Evolution~\cite{cfte} encapsulates temporal trajectories into $4\times4$ matrices, and Latent Image Animation~\cite{lia} extracts weighting coefficients for decomposed motion vectors. However, these methods exploit prior knowledge from video contents under the same scenario, hindering their trained models to generalize to more diverse video contents.

\begin{figure}[t]
    \centering
\subfloat[Video-content-prior-based.]
{\includegraphics[height=3.4cm]{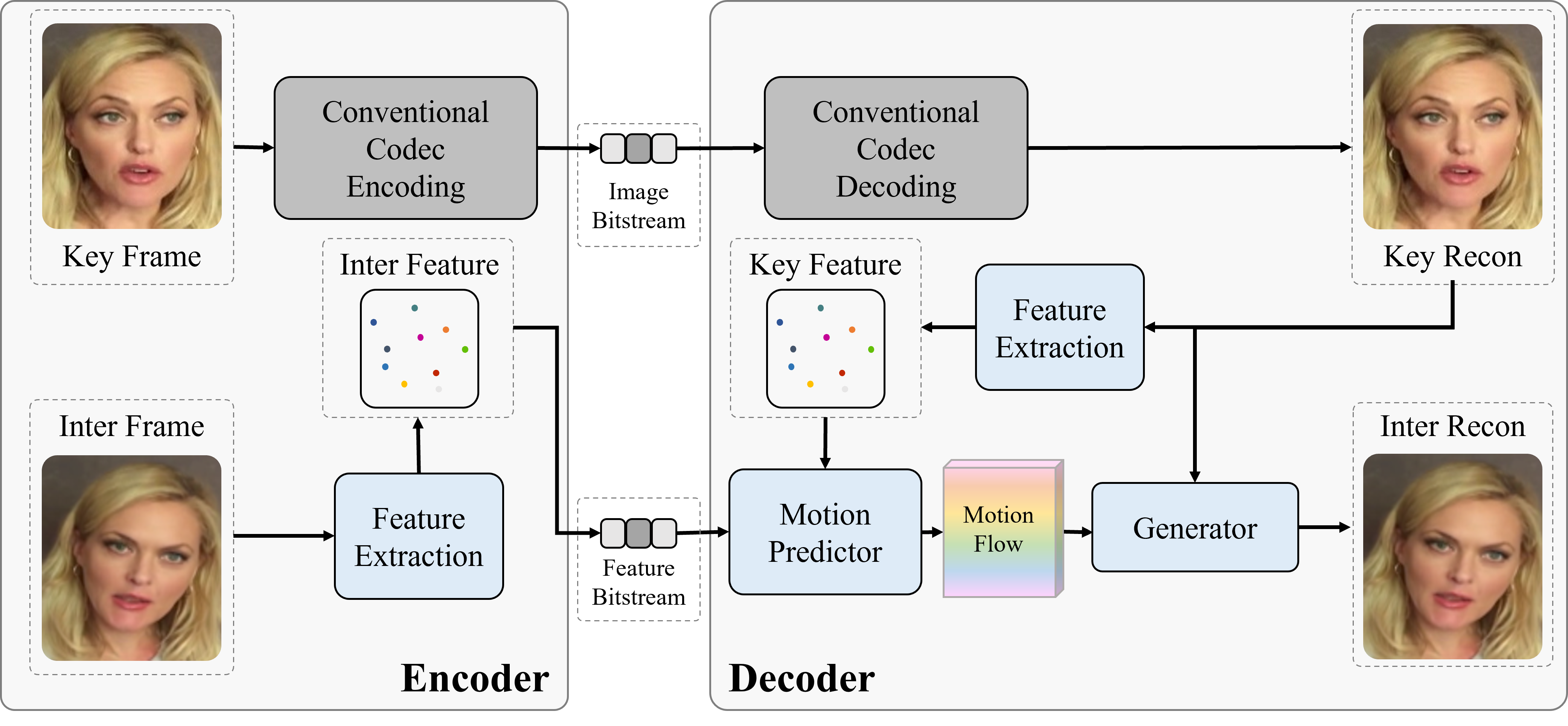}}
\hspace{0.2cm}
\subfloat[Motion-pattern-prior-based.]{\includegraphics[height=3.4cm]{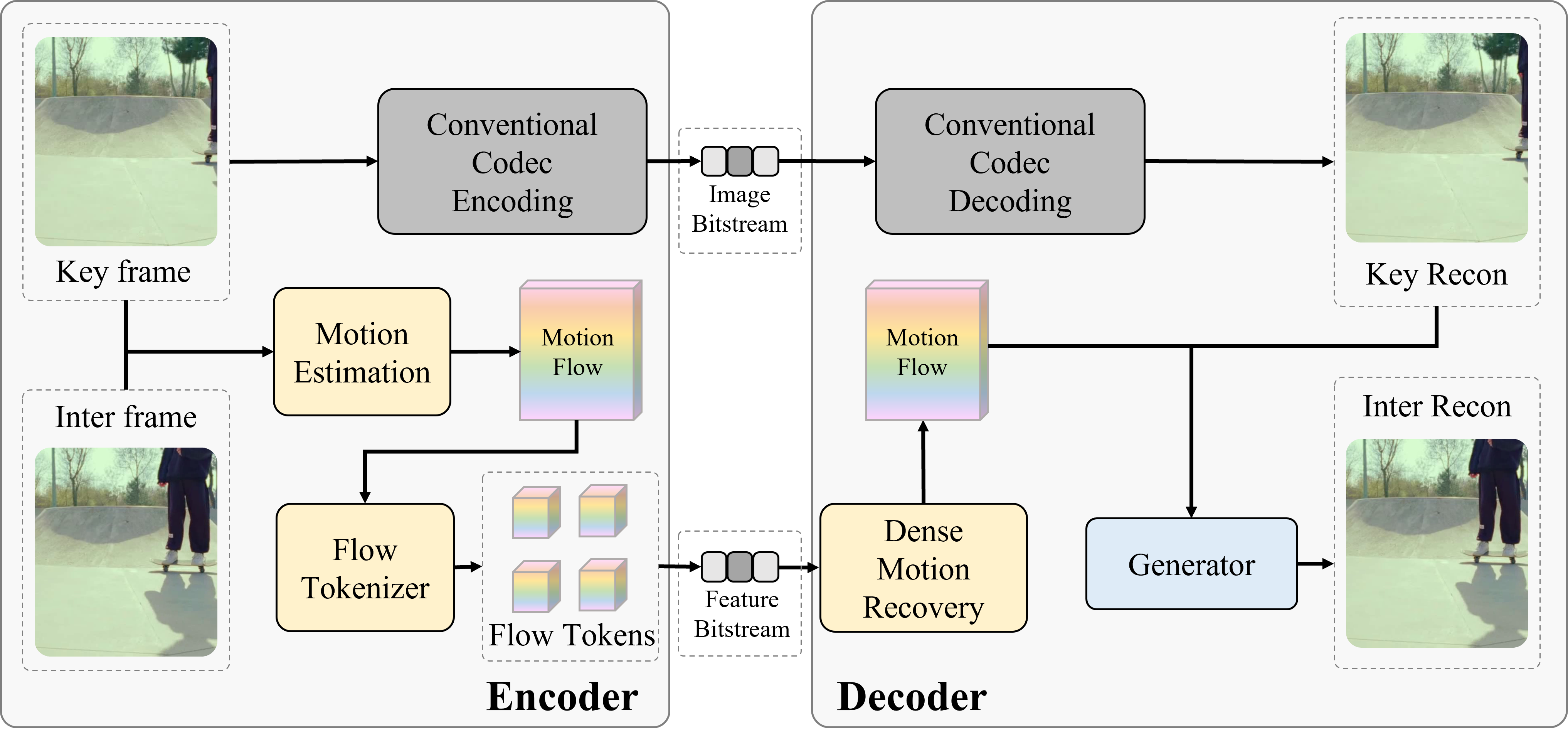}}
\caption{Comparison of the video-content-prior-based and motion-pattern-prior-based generative video coding.}
\label{fig:motivation}

\end{figure}

Recently, Generative Image Dynamics~\cite{generative-image-dynamics} illustrate the effectiveness of learning the distribution of natural motions conditioned on a given image. It can generate natural oscillatory motions across scenes using the Diffusion Model~\cite{diffusion}. Connecting this motion prior with the prior-based generative video coding raises an intriguing question: \textit{Can we characterize specific motion pattern priors into suitable representations that are independent of video contents, so as to enable generative video coding to generalize to a wider range of scenes?} To verify this idea, in this paper, we follow~\cite{generative-image-dynamics, dynamics-dcc} and choose the most common motion patterns, i.e., small motion dynamics in everyday scenes, to explore the possibility of building Generative Scene Dynamics Coding framework~(Dynamics-Codec). Herein, we limit ``small motion dynamics'' to the minor, rigid and non-articulated motion in ubiquitous daily scenes, such as oscillations and rectilinear movements of objects. It is expected that this Dynamics-Codec framework can directly extract compact feature representations from motion flows instead of video contents and realize generative reconstruction for various kinds of dynamic scenes. The difference between the video-content-prior-based and motion-pattern-prior-based generative video coding is illustrated in Figure~\ref{fig:motivation}.

The Dynamics-Codec is developed through two primary efforts. First, a motion tokenizer extracts compact motion tokens from motion flows in a dense-to-sparse manner for efficient compression. To enhance modeling of specific motion patterns, key movement regions are identified through motion sampling, converting dense motions into sparse representations, which are subsequently compressed into compact tokens. Second, a robust decoder reconstructs videos using key frame and motion data across diverse content. Leveraging the pre-trained Stable Video Diffusion (SVD) model~\cite{svd}, trained on a broad data distribution, ensures generalizability across varied scenes. Additionally, a pre-trained Control-Net-like motion flow adaptor~\cite{mofa} enables seamless flow-driven control via feature warping with reconstructed dense motions.

This paper is an extension version of our previous conference paper~\cite{dynamics-dcc}, which only contains \textit{one page} length of preliminary illustration of the motion-pattern-prior-based video coding framework. In this paper, we provide a comprehensive discussion on motivation, related works, detailed technical designs, optimization strategies, and extensive experimental validations.
The contribution of this paper can be summarized as the following,
\begin{itemize}
    \setlength{\parskip}{0.6pt}
    \item We propose a novel generative scene dynamics compression framework that utilizes motion pattern priors instead of video content priors for the ultra-low bit-rate compression and high-quality reconstruction for diverse video contents.
    \item We design an optical flow tokenizer that can transform dense motion between key frame and inter frame into compact representations for extremely low bit-rate compression of temporal information.
    \item We develop a flow-driven decoder that can recover dense motions from the flow tokens and integrate the powerful diffusion-based generator in GVC for visual-pleasing reconstruction across various scenes.
    \item The experiment results show that the proposed Dynamics-Codec can achieve superior rate-distortion performance as well as subjective quality against the state-of-the-art conventional video codec ECM.
\end{itemize}

\section{Related Work}

\subsection{Hybrid Video Compression}

For over forty years, video coding technologies have undergone remarkable advancements, leading to the development of standardized hybrid codecs with outstanding compression performance, such as Advanced Video Coding~(AVC)~\cite{avc}, High Efficiency Video Coding~(HEVC)~\cite{hevc}, and Versatile Video Coding~(VVC)~\cite{vvc}. The Joint Video Experts Team~(JVET), a collaboration between ISO/IEC SC 29 and ITU-T SG16, is currently spearheading efforts to create a next-generation codec that outperforms VVC. This initiative centers on iteratively enhancing coding tools within the Enhanced Compression Model~(ECM) reference software~\cite{ecm}. In parallel, investigations into Neural Network-based Video Coding~(NNVC)~\cite{nnvc} and the refinement of traditional coding techniques~\cite{deep-inloop,deep-intra} aim to push the boundaries of compression efficiency. In this paper, we employ the hybrid codec to encode key frames in video sequences, achieving superior compression efficiency while ensuring high-quality texture references for generating subsequent inter frames.

\subsection{End-to-End Video Compression}

Unlike hybrid video coding, where coding tools are independently designed and optimized, end-to-end coding models are trained holistically in a data-driven manner. Ballé et al. pioneered this approach in image domain with a transform-quantization-coding pipeline using convolutional neural networks and variational autoencoders~\cite{joint,variational,factorized}. Inspired by such philosophy, DVC~\cite{dvc} marked a breakthrough in end-to-end video coding by implementing all components with deep neural networks. Building on DVC~\cite{dvc}, DCVC~\cite{dcvc} introduced conditional coding in the feature domain, while DCVC-TCM~\cite{dcvc-tcm} enhanced compression through temporal context mining. DCVC-HEM~\cite{dcvc-hem} incorporated an efficient spatial-temporal entropy model, and DCVC-DC~\cite{dcvc-dc} increased context diversity across temporal and spatial dimensions. Recently, DCVC-FM~\cite{dcvc-fm} improved quality range and stabilized long prediction chains via feature modulation, outperforming the Enhanced Compression Model (ECM)~\cite{ecm} in the Low-Delay-Bidirectional (LDB) configuration. DCVC-RT~\cite{dcvc-rt} focused on real-time applications, achieving competitive performance with reduced complexity, providing a more practical solution for neural-network-based video coding. 

\subsection{Generative Video Compression with Priors}

To further boost the rate-distortion performances and go beyond the conventional transform-quantization-coding paradigm, generative video compression~(GVC) is proposed to leverage the compact priors of the specific domain and strong generation ability of deep generative models for ultra-low bit-rate coding and perceptually high-quality reconstructions. Specifically, key-frames are compressed by conventional codec while subsequent inter-frames are represented by semantic features. At the decoder side, the key-frames are animated by the motion fields which are derived from the semantic features, to reconstruct inter-frames in a generative manner. Early attempts of GVC mainly focus on evolving the deep image animation model~\cite{fomm} into Generative Face Video Coding~(GFVC)~\cite{gfvc-survey-trans} models by leveraging rich semantic priors of human faces. To achieve that, diverse feature representations are explored including 2D key-points~\cite{dac}, 3D key-points~\cite{fv2v}, compact matrices~\cite{cfte,cttr}. In parallel, hybrid schemes are also incorporated for better reconstruction quality, such as multi-layer coding~\cite{hdac}, predictive coding~\cite{rdac}, multi-frame reference~\cite{mr-dac}, multi-view fusion~\cite{multiview}, bi-directional prediction~\cite{bi-direction}, progressive coding~\cite{progressive-gfvc}, scalable coding~\cite{pleno} and interactive coding~\cite{ifvc}. To develop a more practical GFVC system and collaborate with standardized conventional codecs, JVET has investigated the integration of GFVC techniques into conventional codec with Supplimemtary Enhancement Information~(SEI) messages~\cite{gfvc-sei-trans}.

Subsequently, the GVC framework is extended to more complicated human body contents with articulated movements~\cite{mraa}. Principle-Component-Analysis-based key-points representation is first explored~\cite{mraa-codec}, then MTTF~\cite{mttf} utilizes multi-granularity features to achieve both precise feature transmission and precision motion recovery, IHVC~\cite{ihvc} incorporates interactive semantics for controllable body parts manipulation, and IMT~\cite{imt} leverages implicit motion representation with attention-based motion transfer instead of explicit motion field with warpping-based deformation, whose effectiveness has previously been verified on other contents like flowers and foliage~\cite{imf}. To further expand the dimension of GVC, Sparse2Dense~\cite{sparse2dense} facilitates 3D vertices' prediction from 3D key-points along with human video reconstruction. However, existing GVC methods predominantly focus on video-content-prior-based schemes, which limits their generalizability to other domain's contents, once trained on a specific domain like human faces or human bodies. Furthermore, these existing GVC schemes mainly utilize Generative Adversarial Networks~(GAN)~\cite{gan} for reconstruction, whose generation capability has been suppressed by Diffusion Models~\cite{diffusion} or Auto-Regressive Models~\cite{llm_beates_diffusion} in recent year. In this paper, we attempt to explore motion-pattern-prior-based GVC framework that can generalize to various scene dynamics contents, and leverage the powerful diffusion-based generator for high-quality reconstruction.

\subsection{Motion-Driven Image-to-Video Generation}

Temporal modeling is an essential for in video generation, and many works attempt to incorporate motion field as an explicit driving condition. Early attempts include animating fluid elements using dense motion predicted from source image and optional arrow-like sparse motion~\cite{controllale_fluid,animate_fluid}.
More recently, the development of Video Diffusion Models~\cite{lvdm} has significantly advanced the image-to-video generation field with longer sequence length, higher resolution, better temporal coherence and more diverse controlling signals~\cite{diffusion-review}. As for motion-driven scenarios, DrugNUWA~\cite{dragnuwa} samples sparse trajectories from dense motion and integrate them with source image and text hints into diffusion generator with multi-scale fusion, Motion-I2V~\cite{motion-i2v} utilizes two-stage pipeline of text-guided motion generation and motion-guided video generation, and MotionCtrl~\cite{motionctrl} decomposes camera motion and objective motion and offers independent control. To further improve the accuracy of temporal alignment, MOFA~\cite{mofa} designs a motion adaptor to warp multi-scale image features with dense motion, Tora~\cite{tora} introduces temporal attention mechanism to integrate motion patches into Diffusion Transformer~(DiT), while DragAnything~\cite{draganything} achieves objects-level control with an entity semantic representation. In this paper, we propose to leverage the powerful generation ability of pre-trained motion-driven diffusion model for scene dynamics reconstruction in GVC framework, which makes it possible to explore the prior modelling from motion patterns instead of video contents to enhance the generalizability of GVC between various scenes.

\section{Proposed Method}

\begin{figure}[t]
    \centering
\includegraphics[width=15cm]{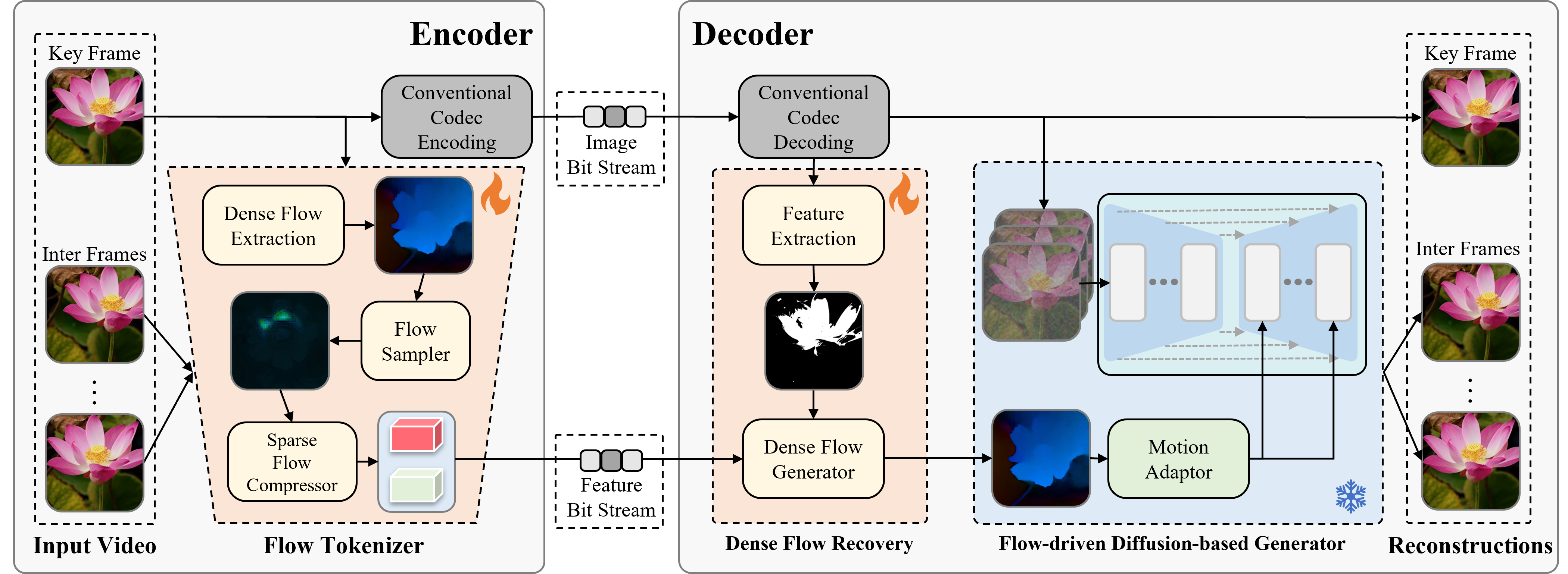}
\caption{The overall framework of proposed Dynamics-Codec. The networks in orange blocks are optimized from scratch, and the networks in blue block is pre-trained.}
    \label{fig:overall}

\end{figure}

\subsection{The Dynamics-Codec Framework}

The detailed structure of proposed Dynamics-Codec is shown in Figure~\ref{fig:overall}. At the encoder side, the key frame~(i.e., the first frame of the video sequences) is compressed by conventional codec and transmitted as image bit-stream. For inter frames, dense optical flows are first extracted between the key frame and every subsequent inter frame. Then, each dense flow is sampled by a watershed-based method~\cite{cmp,watershed} to obtain the sparse motion and the corresponding motion mask, which are further down-sampled and vectorized to form motion tokens. To further eliminate the coding redundancy, all tokens are inter-predicted with the adjacent token, and the quantized residuals are encoded by Context Adaptive Binary Arithmetic Coding~(CABAC).

At the decoder side, the reconstructed key frame is decoded by conventional codec from the image bit-stream, and fed into a feature extractor.
Meanwhile, the motion tokens are obtained by context-based entropy decoding and feature compensation from the feature
bit-stream. The dense motions are subsequently reconstructed by leveraging the internal relationship between the extracted key frame features and the decoded motion tokens. Finally, the recovered dense motions are fed into a flow-driven diffusion-based generator for denoising generation of reconstructed inter frames.

\subsection{Optical Flow Tokenizer}

\begin{figure}[t]
    \centering
\includegraphics[width=15cm]{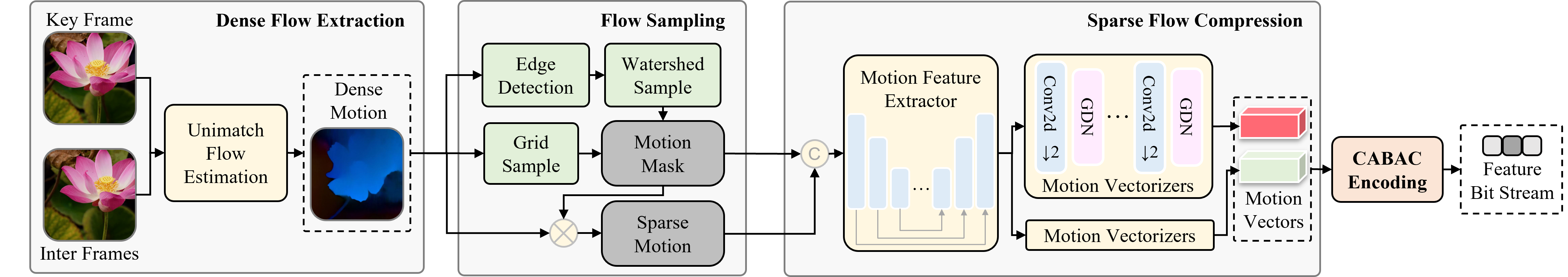}
\caption{The detailed structure of motion tokenizer.}
    \label{fig:tokenizer}

\end{figure}

Feature representation is essential to generative video coding. Previous GVC works mainly focus on utilizing video content priors for characterizing temporal trajectory on homogeneous data such as human face or human body~\cite{generativevisualcompressionreview}. In this paper, we advance the idea of video-content prior to motion-pattern prior that requires direct feature extraction from optical flows. To preserve the major motion information and benefit the efficient representation, the motion tokenizer is designed as a dense-to-sparse scheme, as shown in Fig.~\ref{fig:tokenizer}.

Specifically, dense motion is first extracted with an off-the-shelf and latest motion estimation model Unimatch~\cite{unimatch}, which utilizes a transformer-based multi-task framework to solve the unified dense correspondence matching problem.
Without loss of generality, we denote the key frame and inter frame as $\textbf{I}$ and $\textbf{P}$ with the dimension of $3\times H \times W$, the motion extraction can be denoted as,
\begin{equation}
    \textbf{f}_{d} = \Theta (\textbf{I}, \textbf{P})
\end{equation}
where $\Theta$ denotes motion prediction and $\textbf{f}_{d} \in \mathbb{R}^{H \times W \times 2} $ denotes the motion field in the format of coordinate grid.

To select the primary motion trajectories under the consideration of both its appearance and intensity, the flow edge is first extracted from the dense flow to distinguish the boundary of moving regions, then a topological-distance watershed map is created~\cite{watershed} to sample sparse flow in the center areas of moving regions, finally Non-maximum Suppression~\cite{canny} is leveraged to obtain the key-points and corresponding motion mask~\cite{cmp}. Meanwhile, a uniform grid-sample is also utilized to ensure the coverage of the entire motion field. In this way, the dense motion is transformed to sparse motion and motion mask, 
\begin{equation}
  \textbf{f}_{s}, \textbf{m}_{s} = \Phi (\textbf{f}_{d})
\end{equation}
where $\Phi$ denotes motion sampling process. $\textbf{f}_{s} \in \mathbb{R}^{H \times W \times 2} $ and $\textbf{m}_{s} \in \mathbb{R}^{H \times W \times 1} $ denote the sparse motion and the motion mask respectively.

Then, a sparse flow compression is performed to transform sparse motions to compact motion tokens, where the motion feature is first obtained by a flow feature extractor $\textbf{e}_{f}$,
\begin{equation}
    \textbf{y}_{f} = \textbf{e}_{f}(concat[\textbf{f}_{s}, \textbf{m}_{s}])
\end{equation}
where $\textbf{y}_{f}$ denotes the flow feature and $concat$ denotes the concatenation operation. Herein, this flow feature extractor is designed as a U-Net~\cite{unet} like structure with up-sampling, down-sampling and short-cut connections. Specifically, five down-sample layers are followed by five symmetrically designed up-sample layers. Each layer has a rescaling factor of 2 and the outputs of each down-sample layer are concatenated to the corresponding up-sampling layer.
Then, two motion token vectors are derived by two vectorizers $\textbf{v}_{w}$ and $\textbf{v}_{b}$,
\begin{equation}
    \textbf{w} = \textbf{v}_{w}(\textbf{y}_{f})
\end{equation}
\begin{equation}
    \textbf{b} = \textbf{v}_{b}(\textbf{y}_{f})
\end{equation}
where $\textbf{w} \in \mathbb{R}^{N_{v} \times 1}$ and $\textbf{b} \in \mathbb{R}^{N_{v} \times 1}$. $N_{v}$ denotes the number of tokens in each vector. Here, the vectorizers are designed as cascaded convolutional layers and Generalized Divisive Normalization layers~\cite{factorized}. Specifically, seven convolution layers with down-sample factor of 2 and one convolutional layers with down-sample factor of 3 are used to transform the spatial size of 384 to 1. In this way, high-fidelity dense motions are transformed to very compact motion tokens, which served as high-level semantic motion hints, through the guidance of intermediate sparse motions. Such a process could exploit the compact motion pattern prior in motion flows, benefiting both high compressibility of transmitted features and decent accuracy of subsequent motion recovery.

\subsection{Dense Motion Recovery}
\begin{figure}[t]
    \centering
\includegraphics[width=7cm]{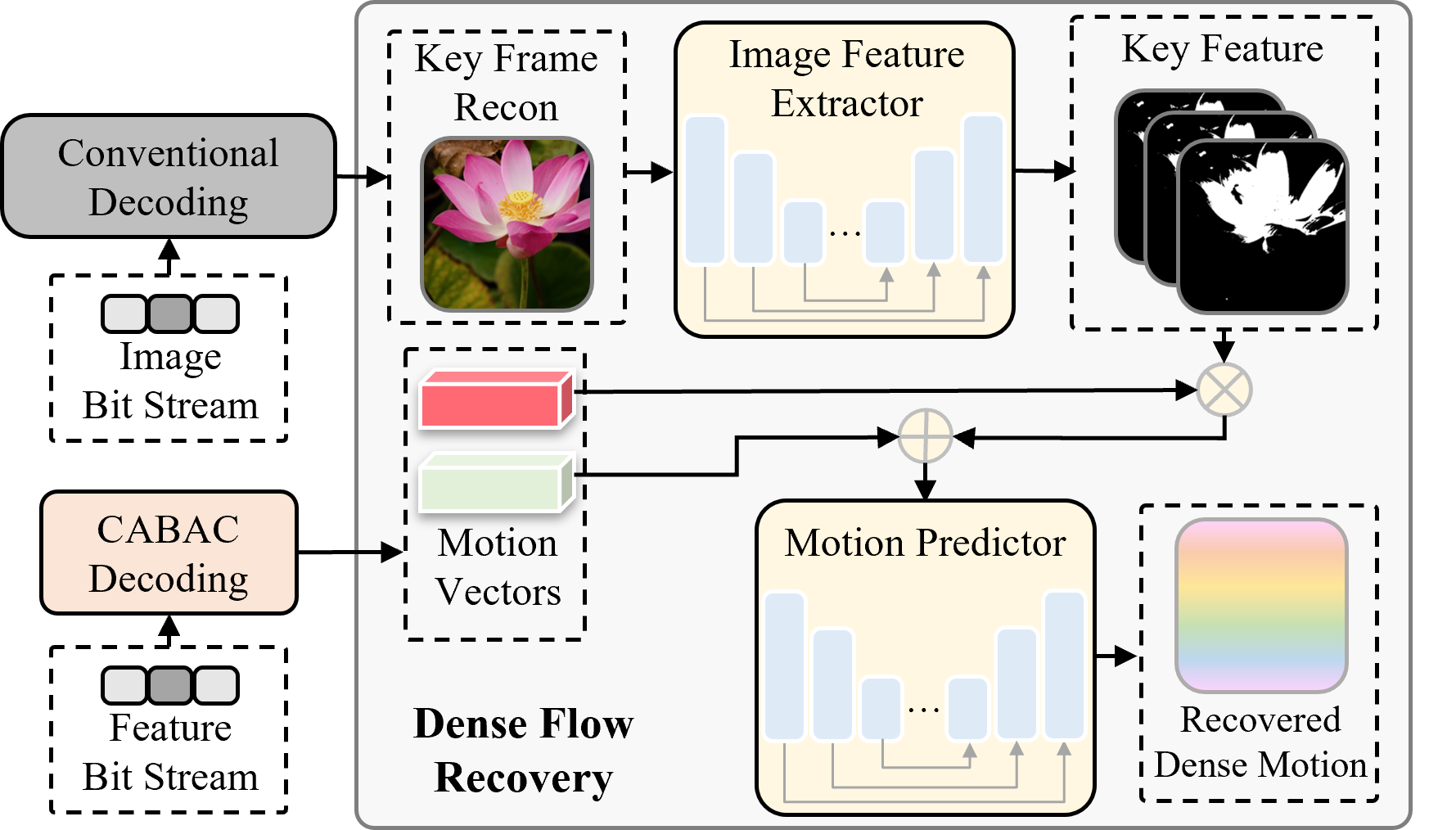}
\caption{The detailed structure of dense flow recovery module.}
    \label{fig:recovery}
\end{figure}

At the decoder side, the dense motion is recovered from the decoded motion tokens with the assistance of reconstructed key frame by a dense motion recovery module, as shown in Fig.~\ref{fig:recovery}.
Specifically, the reconstructed key frame is denoted as $\hat{\textbf{I}}$, and it is first fed into an image feature extractor $\textbf{e}_{i}$,
\begin{equation}
    \textbf{y}_{i} = \textbf{e}_{i}(\hat{\textbf{I}})
\end{equation}
where $\textbf{y}_{i}\in \mathbb{R}^{N_{v} \times H \times W}$ denotes the image feature. Here, the image feature extractor is designed as a U-Net like structure, where five down-sample layers are followed by five symmetrically designed up-sample layers, and each layer has a rescaling factor of 2 and the outputs of each down-sample layer are concatenated to the corresponding up-sampling layer. The decoded motion token vector $\hat{\textbf{w}}$, $\hat{\textbf{b}}$ and the image feature are processed by a motion predictor $\textbf{g}_{f}$,
\begin{equation}
    \hat{\textbf{m}}_{f} = \textbf{g}_{f}(\textbf{y}_{i} \cdot \hat{\textbf{w}} + \hat{\textbf{b}})
\end{equation}
where $\hat{\textbf{m}}_{f}\in \mathbb{R}^{H \times W \times 2}$ denotes the recovered dense motion flow and ``$\cdot$'' denotes channel-wise multiplication. Here, the flow generator is designed with the same structure as the image feature $\textbf{e}_{i}$. 

\subsection{Flow-driven Generator}
In previous GVC schemes, generators are usually crafted following the structures of Generative Adversarial Network~\cite{gan} or Variational Auto-Encoder~\cite{vae}. The resulting GVC methods can only address single scenario application with one trained model, limiting their generalizability to more diverse contents.
Recently, the Diffusion Model~\cite{diffusion} is proposed to perform generation with the denoising process and exhibits the outstanding capability of learning complex data distribution. It can be integrated with versatile conditions for tailored generative vision tasks~\cite{diffusion-review}. Motion trajectory is also leveraged as the control signal for interactive image-to-video generation~\cite{motion-i2v,motionctrl,mofa,dragnuwa,tora}. 
However, they are not yet explored in GVC framework.
In this paper, a pre-trained SVD~\cite{svd} and motion adaptor~\cite{mofa} are utilized to ensure the robustness of motion-driven generation and its generalizability across various contents.

Following the practice in~\cite{mofa}, the recovered dense motion $\hat{\textbf{m}}_{f}$ is fed into the motion-adaptor and used to warp the multi-scale features of reconstructed key frame, which are further fused with SVD encoder features. Specifically, a pre-trained SVD encoder $\textbf{E}_{ref}$ is used as reference encoder to extract the key frame features. Then, the features are warped by the recovered dense motion $\hat{\textbf{m}}_{f}$. The warped features are further fused to a trainable SVD encoder $\textbf{E}_{fus}$, which takes the noisy key frame as input and its features are finally served as denoising condition for the SVD decoder. The process can be formulated as,
\begin{equation}
    \textbf{c} = \textbf{E}_{fus}(\hat{\textbf{I}}+\textbf{n}, Warp(\hat{\textbf{m}}_{f}, \textbf{E}_{ref}(\hat{\textbf{I}})))
\end{equation}
where $\textbf{n}$ denotes the Gaussian noise added to the key frame, $Warp(\cdot)$ denotes the warping operation, and $\textbf{c}$ denotes the denoising condition.
The subsequently inter frame $\hat{\textbf{P}}$ generation can be completed by the pre-trained SVD decoder $\textbf{D}_{svd}$ with the trajectory guidance from $\textbf{c}$,
\begin{equation}
    \hat{\textbf{P}} = \textbf{D}_{svd}(\hat{\textbf{I}}+\textbf{n}, \textbf{c})
\end{equation}
In this way, massive training data can be introduced by pre-trained models without any additional bit-rate or training cost, which can consequently guarantee the visual-pleasing reconstruction of the flow-driven decoder.

\subsection{Optimization}
To bridge the gap between proposed motion token representations and pre-trained flow-driven diffusion-based generator, the optimization of Dynamics-Codec is aimed to align recovered dense motions with the original dense motion inputs of the motion adaptor. To achieve that, only flow tokenizer and dense motion recovery module are optimized, and the original dense motion are provided as supervision signals. Here, the original motion predictor of the motion adaptor is denoted as $\phi$, and L1 loss is used to measure the coordinate-wise error between ground truth flows and generated flows. The optimization objective can be written as,
\begin{equation}
    \mathcal{L} = ||\hat{\textbf{m}}_{f}-\phi(\textbf{I},\textbf{P})||_{1}
\end{equation}
where $||\cdot||_{1}$ denotes the L1 norm.

\section{Validations}

\subsection{Experimental Settings}
\subsubsection{Dataset}
\begin{figure}[t]
  \centering
\subfloat[Motion dynamic test set]{\includegraphics[width=13cm]{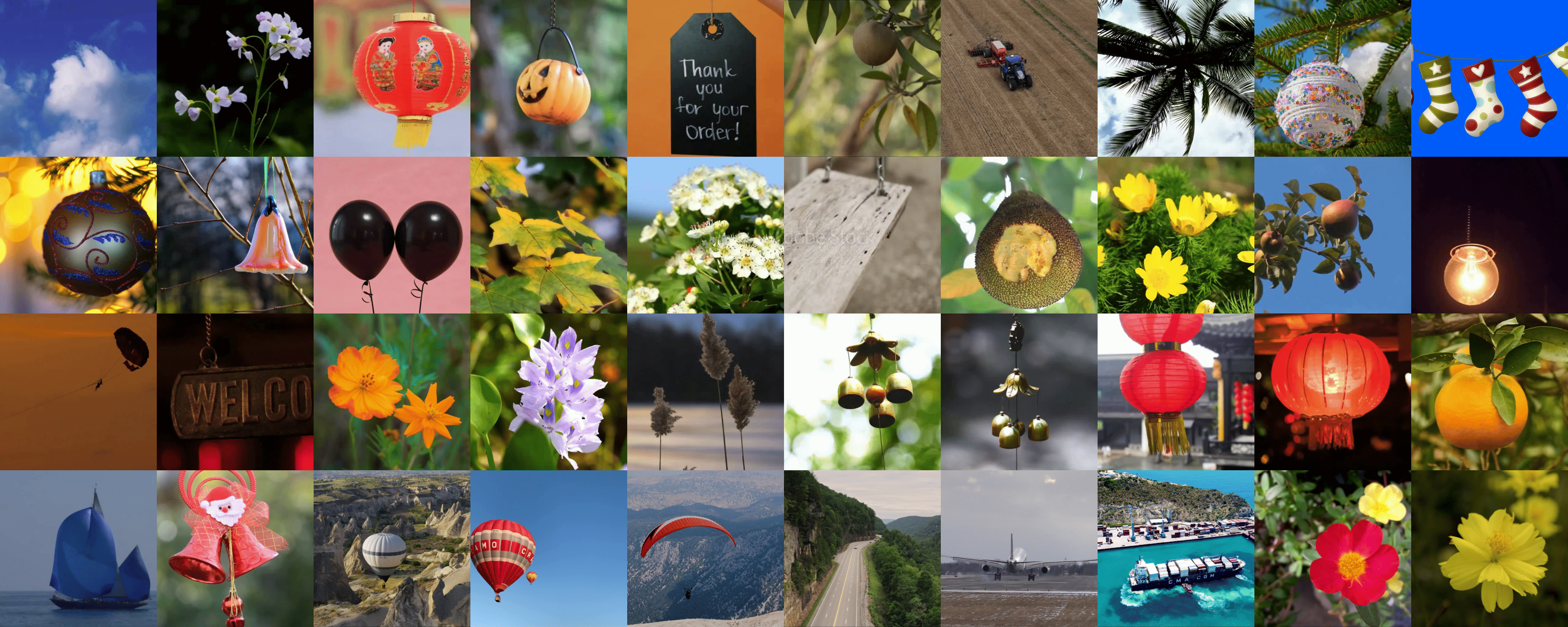}}
\\
\subfloat[JVET-GFV test set]{\includegraphics[width=13cm]{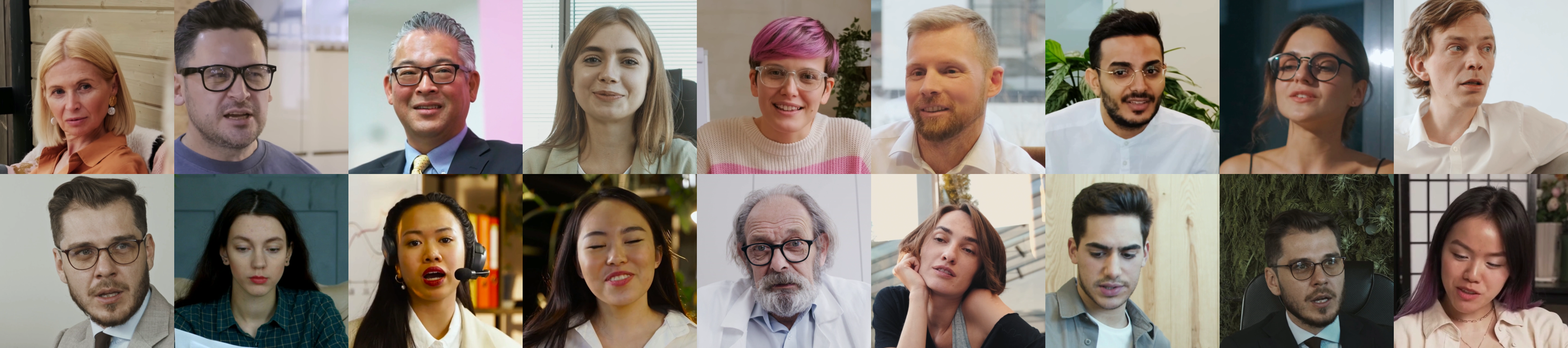}}
\caption{The overview of test sets.}
  \label{fig:testset}
\end{figure}

We collect small motion dynamic videos from the Internet, with various scenes and ranging from size of 720p to 1024p. We square-crop the main areas of scenes dynamics and resize them to resolution of 384 $\times$ 384 for training and evaluation.
For training, total 900 videos are used to extract their ground truth dense motions between the key frame and every inter frame.  
For evaluation, we include two different datasets, as shown in Figure~\ref{fig:testset}, to verify the effectiveness of proposed motion-pattern-based prior and generalizability of proposed method. 
For the small motion dynamic videos~(denoted as motion dynamic test set in the subsequent context), 40 videos with diverse video contents are utilized, and each sequence contains 50 frames with frame rate of 25 fps. We also include talking-face contents for evaluation from JVET-GFV test set~\cite{gfvc-ctc}. We select the first 50 frames of all 18 sequences from Class D, which have head-and-shoulder contents and resolution of 512 $\times$ 512, and resize them to 384 $\times$ 384 to evaluate the performance of proposed method on talking-face contents.

\subsubsection{Implementation Details}
We train the flow tokenizer and dense flow recovery module jointly without entropy coding. The size of image/flow features are set as 384 $\times$ 384, and the number of motion tokens in each token vector is set as 20, yielding totally 40 parameters. We implement the networks with Pytorch framework, which are then optimized by Adam optimizer with $\beta_{1} = 0.5$, $\beta_{1} = 0.999$ and learning rate of $10^{-4}$.  We also use cosine annealing learning rate scheduler with minimum learning rate of $5 \cdot 10^{-7}$.  We train the networks on NVIDIA  GeForce RTX 3090 GPUs for 100 epochs with the batch size of 32.

\subsubsection{Evaluation Settings}
For evaluation metrics, we follow the common practice in generative video coding~\cite{cttr,mttf}, and choose two perceptual measurements that are commonly used for generative contents, i.e., Learned Perceptual Image Patch Similarity~(LPIPS)~\cite{lpips}, Deep Image Structure and Texture Similarity~(DISTS)~\cite{dists}. These two metrics measure the mean square error and structural similarity on feature maps extracted by VGG network and show high correlation with human perception~\cite{gfvc-survey-trans}.  Additionally, we also choose Frechet Video Distance (FVD)~\cite{unterthiner2019fvd} to evaluate the temporal consistency by capturing the temporal dynamics and comparing the feature distribution between the original reconstructed videos.
Natural Image Quality Evaluator~(NIQE)~\cite{niqe}, a commonly used general purpose no-reference metric is also included to evaluate the naturalness of the decoded videos.
\begin{figure}[t]
  \centering
  \subfloat[Rate-DISTS]
  {\includegraphics[width=5.7cm]{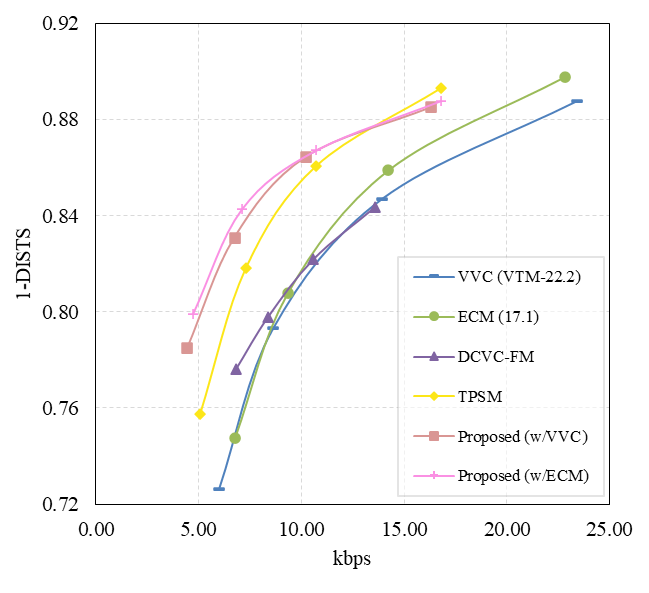}}
  \hspace{0.2cm}
  \subfloat[Rate-LPIPS]{\includegraphics[width=5.7cm]{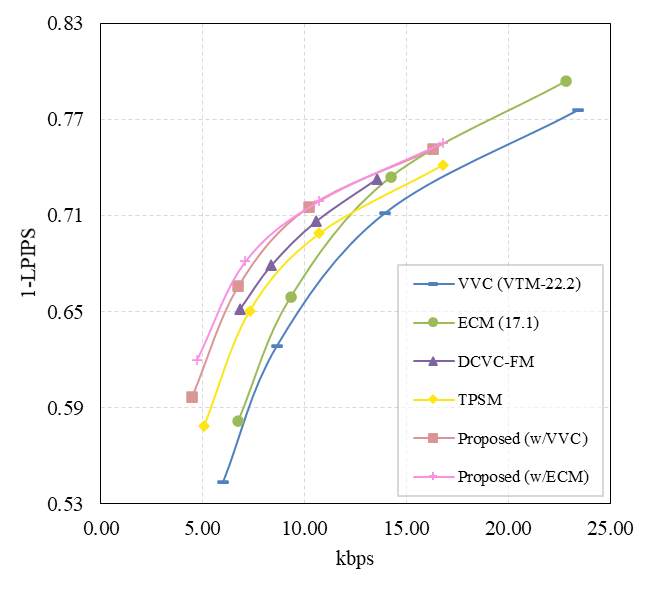}}
  \\
  \subfloat[Rate-FVD]{\includegraphics[width=5.7cm]{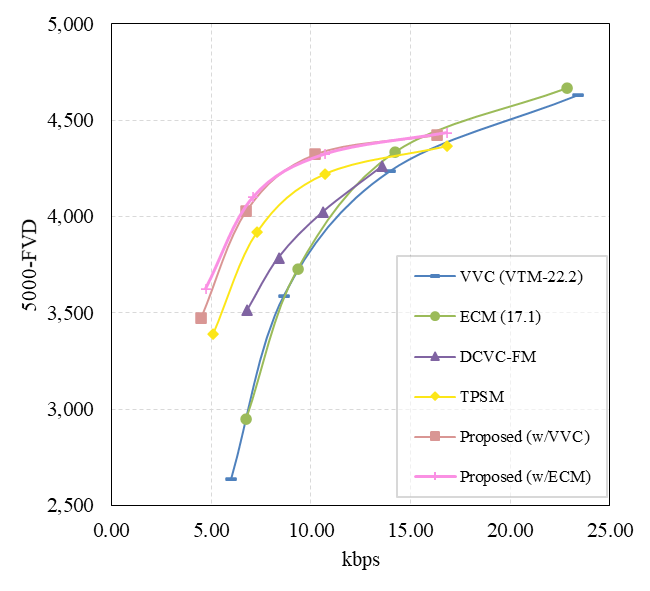}}
  \hspace{0.2cm}
  \subfloat[Rate-NIQE]{\includegraphics[width=5.7cm]{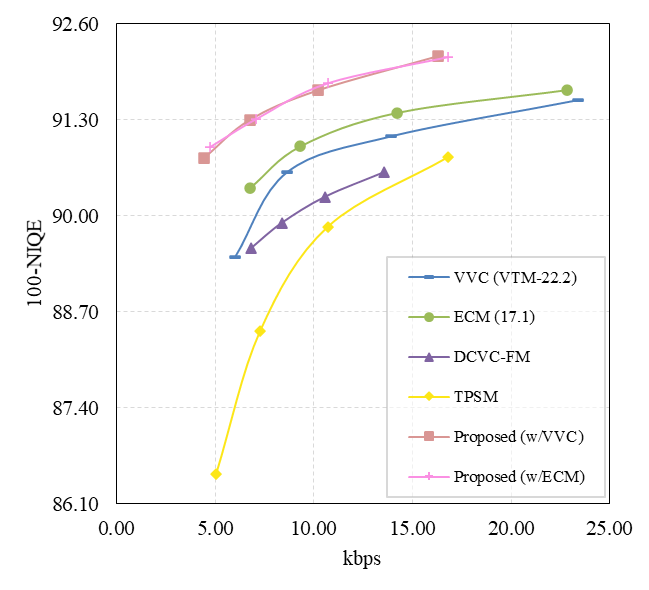}}
   \caption{Rate-distortion performance comparisons with VVC~\cite{vvc}, ECM~\cite{ecm}, DCVC-FM~\cite{dcvc-fm} and TPSM~\cite{tpsm} in terms of DISTS, LPIPS, FVD and NIQE for motion dynamic test set.}
   \label{fig:rd}
  \end{figure}

  \begin{figure}[t]
    \centering
    \subfloat[Rate-DISTS]
    {\includegraphics[width=5.7cm]{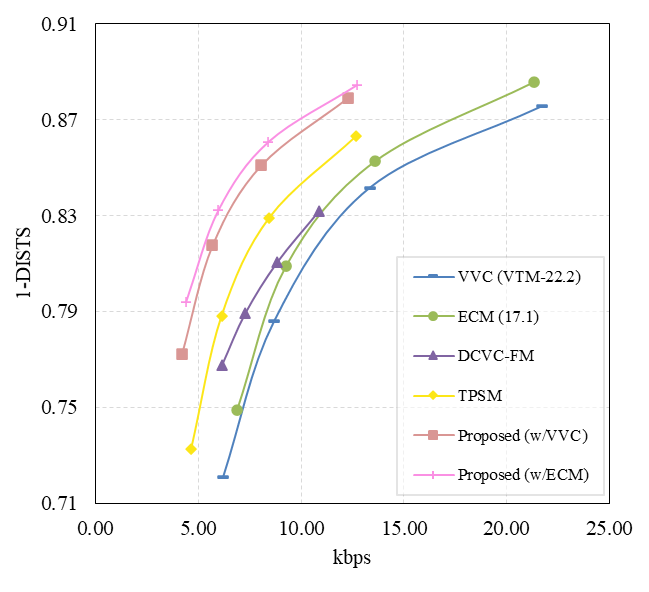}}
    \hspace{0.2cm}
    \subfloat[Rate-LPIPS]{\includegraphics[width=5.7cm]{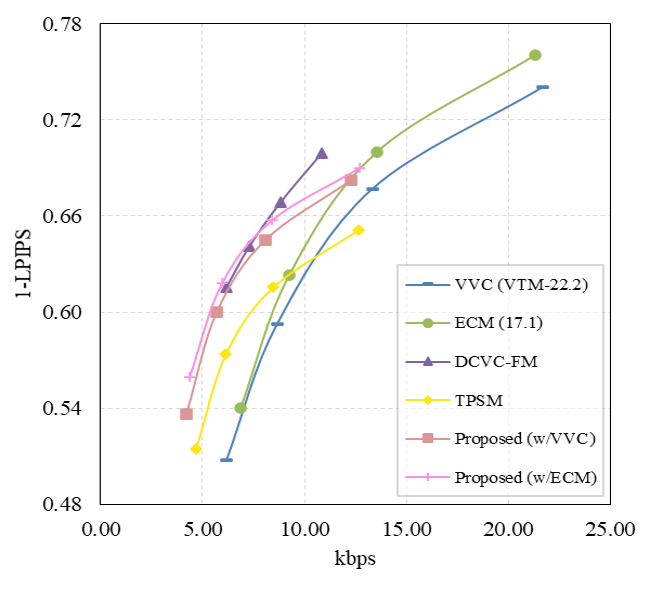}}
    \\
    \subfloat[Rate-FVD]{\includegraphics[width=5.7cm]{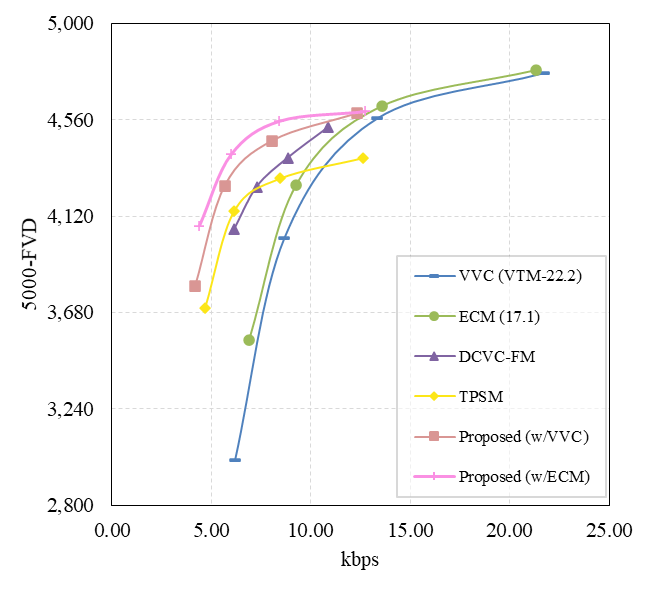}}
    \hspace{0.2cm}
    \subfloat[Rate-NIQE]{\includegraphics[width=5.7cm]{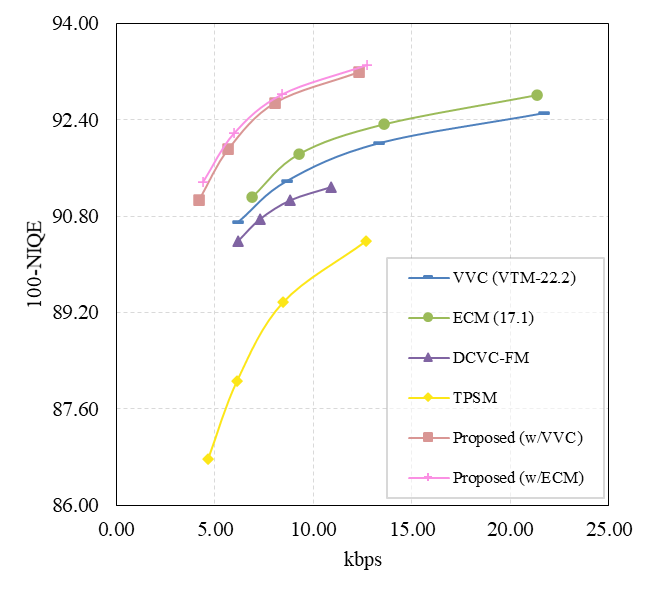}}
     \caption{Rate-distortion performance comparisons with VVC~\cite{vvc}, ECM~\cite{ecm}, DCVC-FM~\cite{dcvc-fm} and TPSM~\cite{tpsm} in terms of DISTS, LPIPS, FVD and NIQE for JVET-GFV~\cite{gfvc-ctc} test set.}
     \label{fig:rd_face}
    \end{figure}
\subsection{Compared Algorithms}

To verify the effectiveness of the proposed method, we
select two state-of-the-art conventional video codec VVC~\cite{vvc}, ECM~\cite{ecm}, one deep-learning-based video codec DCVC-FM~\cite{dcvc-fm},
and one latest generative video codecs TPSM~\cite{tpsm} for comparisons. In the following, we discuss the implementation details.
\begin{itemize}

\item \textbf{Conventional Codec.} VVC~\cite{vvc} is the latest hybrid video coding standard and ECM is the test model of next-generation video standard, which significantly improves the rate distortion performance compared with their predecessors. We adopt VTM 22.2 platform and ECM 17.1 platform with the Low-Delay-Bidirectional (LDB) configuration, where the quantization parameters (QP) are set to 37, 42, 47 and 52.

\item \textbf{Deep-learning-based Video Codec.} DCVC-FM~\cite{dcvc-fm} is one of the latest deep-learning-based video codec that expands the quality range and stabilizes long
prediction chain with feature modulation and outperforms ECM~\cite{ecm}. We implement the DCVC-FM~\cite{dcvc-fm} with the official codebase and set the quantization factors to 0, 5, 10 and 15 for evaluation.

\item \textbf{Generative Video Codec.} TPSM~\cite{tpsm} is one of the state-of-the-art generative codec with thin-plate spline~(TPS) transform for motion estimation. Specifically, it estimates every TPS transform with 5 key-point pairs on each frame and utilizes multi-resolution occlusion masks to improve the generation quality. The key frames are compressed by VVC codec~(VTM 22.2) with QP of 32, 37, 42 and 47.

\item \textbf{The proposed Dynamics-Codec.} For key frame compression, we implement 
both VVC codec with VTM 22.2 platform~(denoted as ``w/VVC'') and ECM 17.1 platform~(denoted as ``w/ECM'') with intra mode and QP of 32, 37, 42, 47. 

\end{itemize}

\subsection{Evaluation Results}
\subsubsection{Rate-Distortion Performance} The RD performances of the proposed Dynamics-Codec and VVC~\cite{vvc} in terms of DISTS, LPIPS, FVD and NIQE on motion dynamic test set are shown in Figure~\ref{fig:rd}. It can be seen that, at the ultra-low bit-rate range from 5 kbps to 15 kbps, the Dynamics-Codec outperforms VVC~\cite{vvc} and ECM~\cite{ecm} among all four metrics on the scene dynamics test set. Compared to DCVC-FM~\cite{dcvc-fm}, Dynamics-Codec shows obviours advantage in terms of DISTS, FVD and NIQE, while performs on par for LPIPS. The specific BD-rate saving against VVC is given in Table~\ref{tab:rd}. Dynamics-Codec can achieve up to 38.80$\%$, 33.86$\%$, 35.13$\%$ and 37.57$\%$ BD-rate saving in terms of Rate-DISTS, Rate-LPIPS, Rate-FVD and Rate-NIQE, which are the highest among all comparison methods.  

For face contents on JVET-GFV~\cite{gfvc-ctc} test set, the RD performances are shown in Figure~\ref{fig:rd_face}. It can be seen that the proposed Dynamics-Codec shows significant advantages over all comparison methods in terms of Rate-DISTS, Rate-FVD and Rate-NIQE, while performs on par with DCVC-FM~\cite{dcvc-fm}. The specific BD-rate saving against VVC~\cite{vvc} is given in Table~\ref{tab:rd_face}. Dynamics-Codec can achieve up to 51.20$\%$, 34.76$\%$, 46.33$\%$ and 58.19$\%$ BD-rate saving in terms of Rate-DISTS, Rate-LPIPS, Rate-FVD and Rate-NIQE, which are the highest among all comparison methods. The results illustrate the effectiveness of the proposed Dynamics-Codec framework. In particular, the internal patterns in motion dynamics are successfully exploited and characterized into compact motion tokens for efficient transmission, and the flow-driven diffusion-based decoder is able to generate high-fidelity inter frames with the assistance of recovered dense motions, across diverse video contents.

\begin{table}[t]
  \caption{RD performance comparisons on scene dynamics test set in terms of average BD-rate savings over the VVC anchor~\cite{vvc}.}
  \label{tab:rd}
  \centering
  \renewcommand\arraystretch{1.2}
  \scalebox{1.0}{
  \begin{tabular}{ccccc}
  \hline
       Algorithm & Rate-DISTS & Rate-LPIPS & Rate-FVD  & Rate-NIQE \\ \hline
        ECM~\cite{ecm} & -1.07\%            & -5.32\%            & 2.36\%   & 0.29\%     \\
        DCVC-FM~\cite{dcvc-fm} & -5.73\%            & -22.31\%            & -3.51\%        & 9.36\% \\
        TPSM~\cite{tpsm}  & -27.26\%            & -17.56\%            & -23.31\%  & 18.08\%       \\
          
         \textbf{Proposed~(w/VVC)} & \textbf{-38.59}\%            & \textbf{-32.54}\%            & \textbf{-35.13}\%  & \textbf{-34.61}\%       \\
         
         \textbf{Proposed~(w/ECM)} & \textbf{-38.80}\%            & \textbf{-33.89}\%            & \textbf{-25.57}\%  & \textbf{-37.57}\%       \\\hline
  \end{tabular}
  }
  \end{table}
  
  \begin{table}[t]
      \caption{RD performance comparisons on JVET-GFV test set in terms of average BD-rate savings over the VVC anchor~\cite{vvc}.}
      \label{tab:rd_face}
      \centering
      \renewcommand\arraystretch{1.2}
      \scalebox{1.0}{
      \begin{tabular}{ccccc}
      \hline
           Algorithm & Rate-DISTS & Rate-LPIPS & Rate-FVD  & Rate-NIQE \\ \hline
            ECM~\cite{ecm} & -7.00\%            & -7.79\%            & -4.13\%   & -3.50\%     \\
            DCVC-FM~\cite{dcvc-fm} & -13.77\%            & -31.03\%            & -12.62\%        & -35.18\% \\
            TPSM~\cite{tpsm}  & -24.57\%            & -15.74\%            & -27.215\%  & -1.44\%       \\
              
             \textbf{Proposed~(w/VVC)} & \textbf{-44.91}\%            & \textbf{-29.57}\%            & \textbf{-34.44}\%  & \textbf{-46.26}\%       \\
             
             \textbf{Proposed~(w/ECM)} & \textbf{-48.50}\%            & \textbf{-32.38}\%            & \textbf{-44.02}\%  & \textbf{-48.38}\%       \\\hline
      \end{tabular}
      }
  \end{table}

  \begin{figure}[t]
    \centering
    \subfloat[Sequence of motion dynamics test set at 6kbps]
    {\includegraphics[width=15cm]{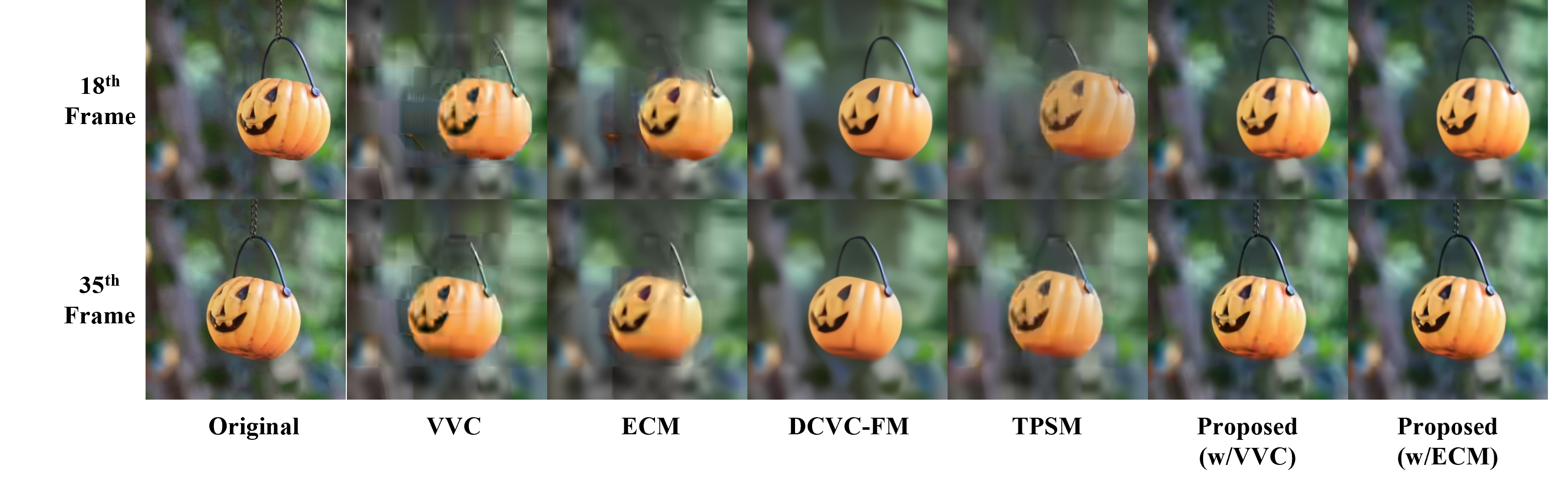}}
    \\
    \subfloat[Sequence of JVET-GFV test set at 10kbps]
    {\includegraphics[width=15cm]{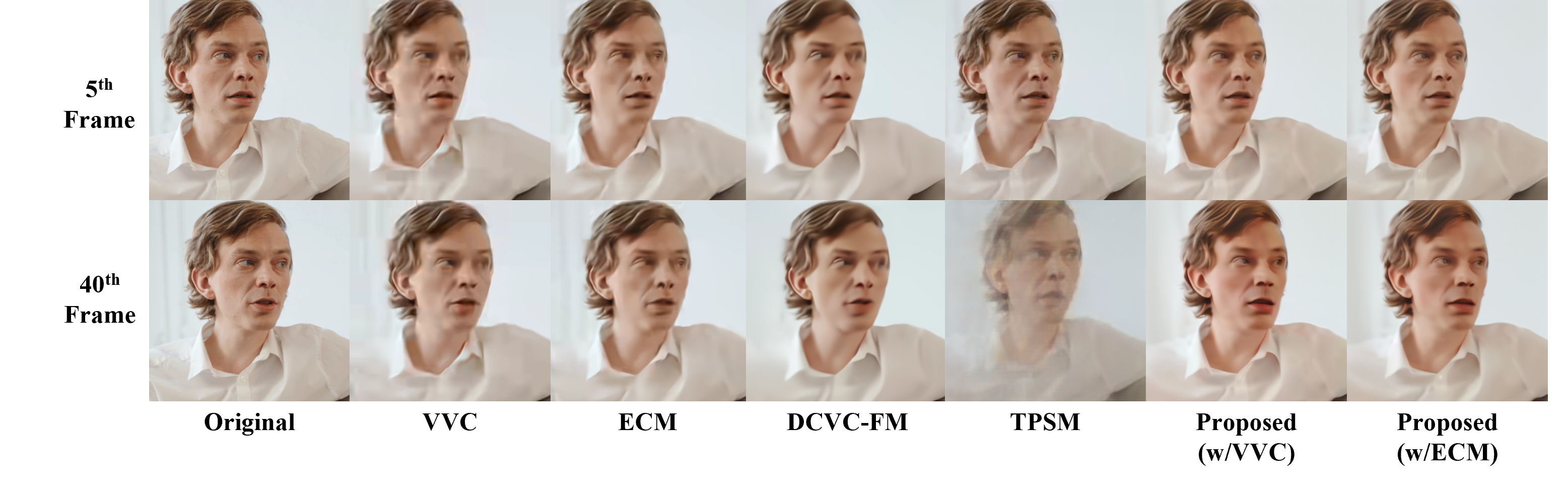}}
   
     \caption{Subective quality comparisons with VVC~\cite{vvc}, ECM~\cite{ecm}, DCVC-FM~\cite{dcvc-fm} and TPSM~\cite{tpsm}. Please refer to the project page for video demos.}
     \label{fig:subj}
    \end{figure}
\subsubsection{Subjective Quality} Subjective visual quality comparisons of proposed Dynamics-Codec and comparison methods are shown in Figure~\ref{fig:subj}. Specifically, reconstructions of motion dynamics sequence at 6kbps and JVET-GFV~\cite{gfvc-ctc} sequence at 10kbps are provided. Two reconstructed frames from each sequence are presented for subjective comparison. It can be observed that, under ultra-low bit-rate, VVC~\cite{vvc} and ECM~\cite{ecm} reconstructions exist annoying blocking effects, DCVC-FM~\cite{dcvc-fm} reconstructions are blurry, and TPSM~\cite{tpsm} reconstructions show obvious deformation with larger objects movements, while the Dynamics-Codec generates more visual-pleasing reconstructions with decent motion accuracy. These results demonstrate the robust generation capability of proposed motion-driven decoder for both high-quality generation and generalizability across diverse scenes. 

Furthermore, we conduct user study to compare our Dynamic-Codec with all other algorithms at similar coding bit-rate. 
Specifically, we choose 15 sequences
and implement ``two alternatives, force choice''~(2AFC) subjective test with 10 participants. During the test, these selected sequences from our method and other compared algorithms are sequentially displayed in a pair-wise manner, and the participants are asked to choose one video from each pair with better quality. To avoid experimental. As shown in Table~\ref{tab:2}, these participants are more inclined to choose our reconstructed videos as the preferred video compared to other reconstruction results. 
In particular, our proposed method shows absolute advantage with more than 90\% preference ratio with VVC~\cite{vvc}, ECM~\cite{ecm} and TPSM~\cite{tpsm}. As for the user preference between DCVC-FM~\cite{dcvc-fm} and ours, our reconstructed videos can still be voted with a higher ratio of 83.33\%. In addition, we also provide average objective results of these tested sequences. At similar ultra-low bit-rate, our method can achieve advantageous objective quality compared with other methods. 

\begin{table}[t]
  \caption{User preference in pairwise comparison in terms of similar coding bits consumption.}
  \label{tab:2}
  \centering
  \renewcommand\arraystretch{1.2}
  \begin{tabular}{ccccccccc}
  \hline
       Comparisons  & kbps& DISTS~($\downarrow$)& LPIPS~($\downarrow$)& FVD~($\downarrow$)& NIQE~($\downarrow$) & User Preference \\ \hline
         VVC~\cite{vvc} / Ours             &  8.04 / 8.03    &0.21 / 0.14  &0.34 / 0.28 &1497.99 / 546.90 &9.67 / 8.68 & 8.00\% / \textbf{92.00}\%   \\
          ECM~\cite{ecm} / Ours            &  8.15 / 8.03   &0.21 / 0.14   &0.33 / 0.28 &1349.76 / 546.90 &9.27 / 8.68 &1.33\% / \textbf{98.67}\%    \\
          DCVC-FM~\cite{dcvc-fm} / Ours    &   8.11 / 8.03   &0.19 / 0.14  &0.28 / 0.28 &1052.31 / 546.90 &10.22 / 8.68 &16.67\% / \textbf{83.33}\%  \\
          TPSM~\cite{tpsm} / Ours          &  8.28 / 8.03   &0.16 / 0.14   &0.30 / 0.28 &960.72 / 546.90 &11.79 / 8.68 & 6.00\% / \textbf{94.00}\%  \\\hline
  \end{tabular}
  \end{table}

\subsubsection{Pixel-level Quality}
\begin{figure}[t]
  \centering
  \subfloat[Rate-PSNR of motion dynamic test set]
  {\includegraphics[width=5.6cm]{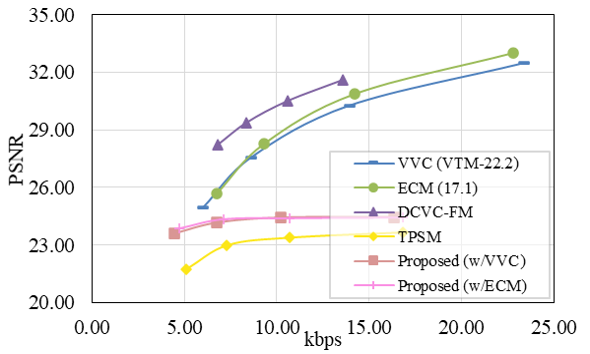}}
  \hspace{0.5cm}
  \subfloat[Rate-SSIM of motion dynamic test set]{\includegraphics[width=5.6cm]{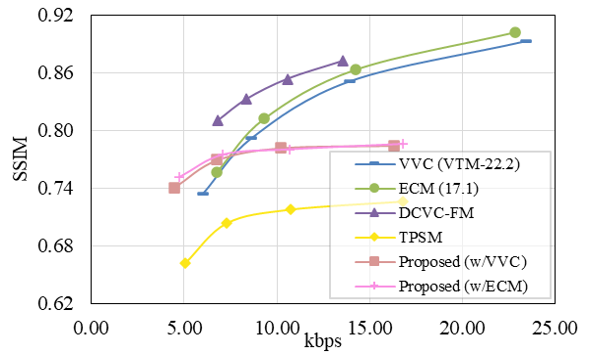}}
  \\
  \subfloat[Rate-PSNR of JVET-GFV test set]{\includegraphics[width=5.6cm]{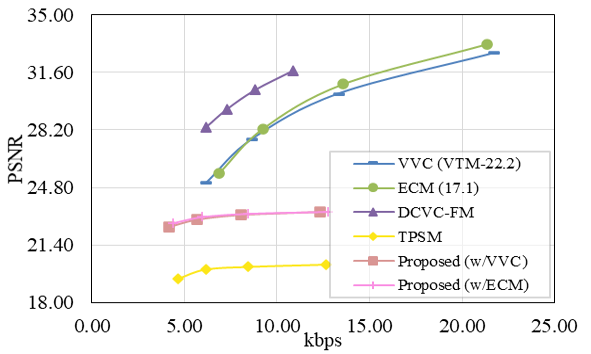}}
  \hspace{0.5cm}
  \subfloat[Rate-SSIM of JVET-GFV test set]{\includegraphics[width=5.6cm]{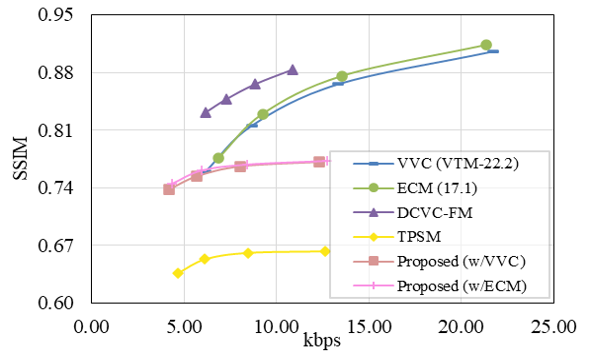}}

   \caption{Pixel-level metrics comparisons with VVC~\cite{vvc}, ECM~\cite{ecm}, DCVC-FM~\cite{dcvc-fm} and TPSM~\cite{tpsm} in terms of PSNR, SSIM.}
   \label{fig:rd_Pixel}
  \end{figure}
The pixel-level quality comparisons in terms of Peak Signal-to-Noise Ratio~(PSNR) and Structural Similarity Index Measure~(SSIM)~\cite{ssim} on two test sets are shown in Fig.~\ref{fig:rd_Pixel}. It can be seen that the proposed Dynamics-Codec performs inferior to VVC~\cite{vvc}, ECM~\cite{ecm} and DCVC-FM~\cite{dcvc-fm} among all test sets, which is mainly due to the generative nature of the proposed method. On the one hand, the generative codecs usually sacrifice pixel-level fidelity for better perceptual quality, which is also verified in previous GVC methods~\cite{cttr,mttf,ifvc}. On the other hand, pixel-level metrics shows lower correlation with human perception under low bit-rate~\cite{gfvc-survey-trans}.
Nevertheless, the proposed Dynamics-Codec can still achieve outperform TPSM~\cite{tpsm} in terms of Rate-PSNR and Rate-SSIM on all test sets, which further demonstrates the powerful generation ability of proposed diffusion-based generator.

\begin{table}[t]
  \caption{Complexity Comparison in terms of encoding time and decoding time~\cite{vvc}.}
  \label{tab:complexity}
  \centering
  \renewcommand\arraystretch{1.2}
  \begin{tabular}{ccccc}
  \hline
Algorithm & Encoding time~(s) & Decoding time~(s) & Total time~(s) \\ \hline
VVC~\cite{ecm} & 1648.16            & 0.73   & 1648.89      \\
ECM~\cite{ecm} & 143.73            & 0.19    & 143.91    \\
DCVC-FM~\cite{dcvc-fm} & 2.75    & 2.60    & 5.35  \\
TPSM~\cite{tpsm}  & 20.69            & 13.81   & 34.57 \\
          
Proposed~(w/VVC) & 26.73 & 117.22  & 143.95         \\
        
Proposed~(w/ECM) & 203.41 & 119.93    & 323.34        \\\hline
  \end{tabular}
\end{table}
\subsection{Complexity Analysis} We measure the encoding time and decoding time of all methods for complexity comparisons. The experiments are conducted with Intel Xeon Silver 4210 CPU @ 2.20GHz and NVIDIA GeForce RTX 3090 GPU. The average time durations to encode and decode four sequences with all different qualities are reported in Table~\ref{tab:complexity}. It can be seen that, the proposed Dynamics-Codec shows competitive encoding time compared with other generative and conventional methods, while the decoding time is relatively high due to the iterative sampling process of diffusion model. In total, the proposed Dynamics-Codec shows largely reduced total time consumption compared to ECM~\cite{ecm} and VVC~\cite{vvc}. Meanwhile, our methods shows higher encoding time when using ECM~\cite{ecm} as the key frame encoder. 

\section{Conclusion}
In this paper, we propose to exploit motion-pattern-prior instead of video-content-prior for generative coding of scene dynamics. With proposed novel Dynamics-Codec, compact motion prior representations are extracted by identifying primary motion components and condensing them into tokens. Meanwhile, robust reconstructions across diverse video contents are ensured by a powerful diffusion-based, motion-driven decoder. The experiment results show that our methods can achieve more than 48\% BD-rate saving against VVC as well as visual-pleasing reconstructions under ultra-low bit-rate on diverse scenes, highlighting better robustness and generalizability of the motion-prior-based generative video coding scheme.

\bibliographystyle{ACM-Reference-Format}
\bibliography{sample-base}

@String{Computing = "Computing" }

@String{Computer = "{IEEE} Computer" }

@String{Springer = "Springer-Verlag" }

@inproceedings{generativevisualcompressionreview,
      title={Generative Visual Compression: A Review}, 
      author={Bolin Chen and Shanzhi Yin and Peilin Chen and Shiqi Wang and Yan Ye},
      year={2024},
      booktitle={IEEE International Conference on Image Processing},
      organization={IEEE}
}

@ARTICLE{vvc,
  author={Bross, Benjamin and Wang, Ye-Kui and Ye, Yan and Liu, Shan and Chen, Jianle and Sullivan, Gary J. and Ohm, Jens-Rainer},
  journal={IEEE Transactions on Circuits and Systems for Video Technology}, 
  title={{Overview of the Versatile Video Coding (VVC) Standard and its Applications}}, 
  year={2021},
  volume={31},
  number={10},
  pages={3736-3764},
  doi={10.1109/TCSVT.2021.3101953}}

@ARTICLE{hevc,
  author={Sullivan, Gary J. and Ohm, Jens-Rainer and Han, Woo-Jin and Wiegand, Thomas},
  journal={IEEE Transactions on Circuits and Systems for Video Technology}, 
  title={{Overview of the High Efficiency Video Coding (HEVC) Standard}}, 
  year={2012},
  volume={22},
  number={12},
  pages={1649-1668},
  doi={10.1109/TCSVT.2012.2221191}}

@inproceedings{lpips,
  title={The unreasonable effectiveness of deep features as a perceptual metric},
  author={Zhang, Richard and Isola, Phillip and Efros, Alexei A and Shechtman, Eli and Wang, Oliver},
  booktitle={Proceedings of the IEEE Conference on Computer Vision and Pattern Recognition},
  pages={586--595},
  year={2018}
}

@article{dists,
  title={Image quality assessment: Unifying structure and texture similarity},
  author={Ding, Keyan and Ma, Kede and Wang, Shiqi and Simoncelli, Eero P},
  journal={IEEE transactions on Pattern Analysis and Machine Intelligence},
  volume={44},
  number={5},
  pages={2567--2581},
  year={2020},
  publisher={IEEE}
}

@INPROCEEDINGS{dac,
  author={Konuko, Goluck and Valenzise, Giuseppe and Lathuilière, Stéphane},
  booktitle={IEEE International Conference on Acoustics, Speech and Signal Processing}, 
  title={Ultra-Low Bitrate Video Conferencing Using Deep Image Animation}, 
  year={2021},
  volume={},
  number={},
  pages={4210-4214},
  doi={10.1109/ICASSP39728.2021.9414731}}

@article{fomm,
  title={First order motion model for image animation},
  author={{A. Siarohin, S. Lathuili`ere, S. Tulyakov, E. Ricci, and N. Sebe}},
  journal={Advances in Neural Information Processing Systems},
  volume={32},
  year={2019}
}

@inproceedings{fv2v,
  title={One-shot free-view neural talking-head synthesis for video conferencing},
  author={Wang, Ting-Chun and Mallya, Arun and Liu, Ming-Yu},
  booktitle={Proceedings of the IEEE/CVF conference on Computer Vision and Pattern Recognition},
  pages={10039--10049},
  year={2021}
}

@INPROCEEDINGS{cfte,
  author={Chen, Bolin and Wang, Zhao and Li, Binzhe and Lin, Rongqun and Wang, Shiqi and Ye, Yan},
  booktitle={Data Compression Conference}, 
  title={Beyond Keypoint Coding: Temporal Evolution Inference with Compact Feature Representation for Talking Face Video Compression}, 
  year={2022},
  volume={},
  number={},
  pages={13-22},
  doi={10.1109/DCC52660.2022.00009}}

@inproceedings{mraa,
  title={Motion representations for articulated animation},
  author={Siarohin, Aliaksandr and Woodford, Oliver J and Ren, Jian and Chai, Menglei and Tulyakov, Sergey},
  booktitle={Proceedings of the IEEE/CVF Conference on Computer Vision and Pattern Recognition},
  pages={13653--13662},
  year={2021}
}

@INPROCEEDINGS{mraa-codec,
  author={Wang, Ruofan and Mao, Qi and Jia, Chuanmin and Wang, Ronggang and Ma, Siwei},
  booktitle={2023 IEEE International Symposium on Circuits and Systems}, 
  title={Extreme Generative Human-Oriented Video Coding via Motion Representation Compression}, 
  year={2023},
  volume={},
  number={},
  pages={1-5},
  doi={10.1109/ISCAS46773.2023.10181664}}

@article{avc,
  title={The {H. 264/AVC} advanced video coding standard: Overview and introduction to the fidelity range extensions},
  author={Sullivan, Gary J and Topiwala, Pankaj N and Luthra, Ajay},
  journal={Applications of Digital Image Processing XXVII},
  volume={5558},
  pages={454--474},
  year={2004},
  publisher={SPIE}
}

@article{ecm,
  title={{JVET-AG0007}: {AHG} report ECM tool
assessment {(AHG7)}},
  author={{X. Li, L. F. Chen, Z. Deng, J. Gan, E. François, H. J. Jhu, X. Li and H. Wang}},
  journal={{The Joint Video Experts Team of ITU-T SG 16 WP 3 and ISO/IEC JTC 1/SC 29, doc. no. JVET-AG0007}},
  year={2024},
  month={January}
}

@article{nnvc,
  title={Designs and Implementations in Neural Network-based Video Coding},
  author={Li, Yue and Li, Junru and Lin, Chaoyi and Zhang, Kai and Zhang, Li and Galpin, Franck and Dumas, Thierry and Wang, Hongtao and Coban, Muhammed and Str{\"o}m, Jacob and others},
  journal={arXiv preprint arXiv:2309.05846},
  year={2023}
}

@article{factorized,
  title={End-to-end optimized image compression},
  author={Ball{\'e}, Johannes and Laparra, Valero and Simoncelli, Eero P},
  journal={arXiv preprint arXiv:1611.01704},
  year={2016}
}

@article{variational,
  title={Variational image compression with a scale hyperprior},
  author={Ball{\'e}, Johannes and Minnen, David and Singh, Saurabh and Hwang, Sung Jin and Johnston, Nick},
  journal={arXiv preprint arXiv:1802.01436},
  year={2018}
}

@inproceedings{joint,
  title={Joint autoregressive and hierarchical priors for learned image compression},
  author={Minnen, David and Ball{\'e}, Johannes and Toderici, George D},
  booktitle={Proceeding of Advances in Neural Information Processing Systems},
  volume={31},
  year={2018}
}

@inproceedings{dvc,
  title={{DVC}: An end-to-end deep video compression framework},
  author={Lu, Guo and Ouyang, Wanli and Xu, Dong and Zhang, Xiaoyun and Cai, Chunlei and Gao, Zhiyong},
  booktitle={Proceedings of the IEEE/CVF Conference on Computer Vision and Pattern Recognition},
  pages={11006--11015},
  year={2019}
}

@inproceedings{dcvc-hem,
  title={Hybrid spatial-temporal entropy modelling for neural video compression},
  author={Li, Jiahao and Li, Bin and Lu, Yan},
  booktitle={Proceedings of the 30th ACM International Conference on Multimedia},
  pages={1503--1511},
  year={2022}
}

@inproceedings{dcvc,
 author = {Li, Jiahao and Li, Bin and Lu, Yan},
 booktitle = {Advances in Neural Information Processing Systems},
 pages = {18114--18125},
 publisher = {Curran Associates, Inc.},
 title = {Deep Contextual Video Compression},
 volume = {34},
 year = {2021}
}

@inproceedings{dcvc-dc,
  title={Neural video compression with diverse contexts},
  author={{Li, Jiahao and Li, Bin and Lu, Yan}},
  booktitle={Proceedings of the IEEE/CVF Conference on Computer Vision and Pattern Recognition},
  pages={22616--22626},
  year={2023}
}

@article{dcvc-tcm,
  title={Temporal context mining for learned video compression},
  author={Sheng, Xihua and Li, Jiahao and Li, Bin and Li, Li and Liu, Dong and Lu, Yan},
  journal={IEEE Transactions on Multimedia},
  volume={25},
  pages={7311--7322},
  year={2022},
  publisher={IEEE}
}

@inproceedings{dcvc-fm,
  title={Neural Video Compression with Feature Modulation},
  author={Li, Jiahao and Li, Bin and Lu, Yan},
  booktitle={{IEEE/CVF} Conference on Computer Vision and Pattern Recognition,
             {CVPR} 2024, Seattle, WA, USA, June 17-21, 2024},
  year={2024}
}

@inproceedings{tpsm,
  title={Thin-plate spline motion model for image animation},
  author={Zhao, Jian and Zhang, Hui},
  booktitle={Proceedings of the IEEE/CVF Conference on Computer Vision and Pattern Recognition},
  pages={3657--3666},
  year={2022}
}

@inproceedings{lia,
title={Latent Image Animator: Learning to Animate Images via Latent Space Navigation},
author={Yaohui Wang and Di Yang and Francois Bremond and Antitza Dantcheva},
booktitle={International Conference on Learning Representations},
year={2022}
}

@ARTICLE{cttr,
  author={Chen, Bolin and Wang, Zhao and Li, Binzhe and Wang, Shiqi and Ye, Yan},
  journal={IEEE Transactions on Circuits and Systems for Video Technology}, 
  title={Compact Temporal Trajectory Representation for Talking Face Video Compression}, 
  year={2023},
  volume={33},
  number={11},
  pages={7009-7023},
  doi={10.1109/TCSVT.2023.3271130}}

@INPROCEEDINGS{hdac,
  author={Konuko, Goluck and Lathuilière, Stéphane and Valenzise, Giuseppe},
  booktitle={2022 IEEE International Conference on Image Processing}, 
  title={A Hybrid Deep Animation Codec for Low-Bitrate Video Conferencing}, 
  year={2022},
  volume={},
  number={},
  pages={1-5},
  doi={10.1109/ICIP46576.2022.10458867}}

@INPROCEEDINGS{rdac,
  author={{Konuko, Goluck and Lathuilière, Stéphane and Valenzise, Giuseppe}},
  booktitle={IEEE International Conference on Image Processing}, 
  title={Predictive Coding for Animation-Based Video Compression}, 
  year={2023},
  volume={},
  number={},
  pages={2810-2814},
  doi={10.1109/ICIP49359.2023.10222205}}

@InProceedings{multiview,
    author    = {Volokitin, Anna and Brugger, Stefan and Benlalah, Ali and Martin, Sebastian and Amberg, Brian and Tschannen, Michael},
    title     = {Neural Face Video Compression Using Multiple Views},
    booktitle = {Proceedings of the IEEE/CVF Conference on Computer Vision and Pattern Recognition Workshops},
    month     = {June},
    year      = {2022},
    pages     = {1738-1742}
}

@INPROCEEDINGS{bi-direction,
  author={Tang, Anni and Huang, Yan and Ling, Jun and Zhang, Zhiyu and Zhang, Yiwei and Xie, Rong and Song, Li},
  booktitle={IEEE International Conference on Multimedia and Expo}, 
  title={Generative Compression for Face Video: A Hybrid Scheme}, 
  year={2022},
  volume={},
  number={},
  pages={1-6},
  doi={10.1109/ICME52920.2022.9859867}}

@misc{ifvc,
      title={Interactive Face Video Coding: A Generative Compression Framework}, 
      author={Bolin Chen and Zhao Wang and Binzhe Li and Shurun Wang and Shiqi Wang and Yan Ye},
      year={2023},
      eprint={2302.09919},
      archivePrefix={arXiv},
      primaryClass={cs.CV},
      url={https://arxiv.org/abs/2302.09919}, 
}

@INPROCEEDINGS{gfvc-review,
  author={Chen, Bolin and Chen, Jie and Wang, Shiqi and Ye, Yan},
  booktitle={2024 Data Compression Conference}, 
  title={Generative Face Video Coding Techniques and Standardization Efforts: A Review}, 
  year={2024},
  volume={},
  number={},
  pages={103-112},
  doi={10.1109/DCC58796.2024.00018}}

@article{gfvc-ctc,
  title={Test conditions and evaluation procedures for generative face video coding},
  author={S. McCarthy and B. Chen},
  journal={{The Joint Video Experts Team of ITU-T SG 16 WP 3 and ISO/IEC JTC 1/SC 29, doc. no. JVET-AJ2035}},
  year={2024},
  month={November}
}

@INPROCEEDINGS{deep-inloop,
  author={Yang, Ruiying and Santamaria, Maria and Cricri, Francesco and Zhang, Honglei and Lainema, Jani and Youvalari, Ramin G. and Hannuksela, Miska M. and Elomaa, Tapio},
  booktitle={2023 IEEE International Conference on Visual Communications and Image Processing}, 
  title={Overfitting NN loop-filters in video coding}, 
  year={2023},
  volume={},
  number={},
  pages={1-5},
  doi={10.1109/VCIP59821.2023.10402710}}

@ARTICLE{deep-intra,
  author={Zhu, Linwei and Kwong, Sam and Zhang, Yun and Wang, Shiqi and Wang, Xu},
  journal={IEEE Transactions on Multimedia}, 
  title={Generative Adversarial Network-Based Intra Prediction for Video Coding}, 
  year={2020},
  volume={22},
  number={1},
  pages={45-58},
  doi={10.1109/TMM.2019.2924591}}

@inproceedings{unet,
  title={U-net: Convolutional networks for biomedical image segmentation},
  author={Ronneberger, Olaf and Fischer, Philipp and Brox, Thomas},
  booktitle={Medical Image Computing and Computer-assisted Intervention--MICCAI 2015: 18th international conference},
  pages={234--241},
  year={2015},
  organization={Springer}
}

@inproceedings{unterthiner2019fvd,
  title={FVD: A new metric for video generation},
  author={Unterthiner, Thomas and van Steenkiste, Sjoerd and Kurach, Karol and Marinier, Rapha{\"e}l and Michalski, Marcin and Gelly, Sylvain},
    booktitle={International Conference on Learning Representations},
  year={2019}
}

@InProceedings{generative-image-dynamics,
    author    = {Li, Zhengqi and Tucker, Richard and Snavely, Noah and Holynski, Aleksander},
    title     = {Generative Image Dynamics},
    booktitle = {Proceedings of the IEEE/CVF Conference on Computer Vision and Pattern Recognition},
    month     = {June},
    year      = {2024},
    pages     = {24142-24153}
}

@article{diffusion,
  title={Diffusion models beat gans on image synthesis},
  author={Dhariwal, Prafulla and Nichol, Alexander},
  journal={Advances in neural information processing systems},
  volume={34},
  pages={8780--8794},
  year={2021}
}

@inproceedings{cmp,
  title={Self-supervised learning via conditional motion propagation},
  author={Zhan, Xiaohang and Pan, Xingang and Liu, Ziwei and Lin, Dahua and Loy, Chen Change},
  booktitle={Proceedings of the IEEE/CVF Conference on Computer Vision and Pattern Recognition},
  pages={1881--1889},
  year={2019}
}

@inproceedings{mofa,
  title={MOFA-Video: Controllable Image Animation via Generative Motion Field Adaptions in Frozen Image-to-Video Diffusion Model},
  author={Niu, Muyao and Cun, Xiaodong and Wang, Xintao and Zhang, Yong and Shan, Ying and Zheng, Yinqiang},
  booktitle={European Conference on Computer Vision},
  year={2024}
}

@article{svd,
  title={Stable video diffusion: Scaling latent video diffusion models to large datasets},
  author={Blattmann, Andreas and Dockhorn, Tim and Kulal, Sumith and Mendelevitch, Daniel and Kilian, Maciej and Lorenz, Dominik and Levi, Yam and English, Zion and Voleti, Vikram and Letts, Adam and others},
  journal={arXiv preprint arXiv:2311.15127},
  year={2023}
}

@inproceedings{watershed,
  title={Use of watersheds in contour detection},
  author={Beucher, Serge},
  booktitle={Proc. Int. Workshop on Image Processing, Sept. 1979},
  pages={17--21},
  year={1979}
}

@article{unimatch,
  title={Unifying flow, stereo and depth estimation},
  author={Xu, Haofei and Zhang, Jing and Cai, Jianfei and Rezatofighi, Hamid and Yu, Fisher and Tao, Dacheng and Geiger, Andreas},
  journal={IEEE Transactions on Pattern Analysis and Machine Intelligence},
  year={2023},
  publisher={IEEE}
}

@article{canny,
  title={A computational approach to edge detection},
  author={Canny, John},
  journal={IEEE Transactions on pattern analysis and machine intelligence},
  number={6},
  pages={679--698},
  year={1986},
  publisher={Ieee}
}

@article{gan,
  title={Generative adversarial networks},
  author={Goodfellow, Ian and Pouget-Abadie, Jean and Mirza, Mehdi and Xu, Bing and Warde-Farley, David and Ozair, Sherjil and Courville, Aaron and Bengio, Yoshua},
  journal={Communications of the ACM},
  volume={63},
  number={11},
  pages={139--144},
  year={2020},
  publisher={ACM New York, NY, USA}
}

@article{vae,
  title={Auto-encoding variational bayes},
  author={Kingma, Diederik P},
  journal={arXiv preprint arXiv:1312.6114},
  year={2013}
}

@ARTICLE{diffusion-review,
  author={Croitoru, Florinel-Alin and Hondru, Vlad and Ionescu, Radu Tudor and Shah, Mubarak},
  journal={IEEE Transactions on Pattern Analysis and Machine Intelligence}, 
  title={Diffusion Models in Vision: A Survey}, 
  year={2023},
  volume={45},
  number={9},
  pages={10850-10869},
  doi={10.1109/TPAMI.2023.3261988}}

@article{dragnuwa,
  title={Dragnuwa: Fine-grained control in video generation by integrating text, image, and trajectory},
  author={Yin, Shengming and Wu, Chenfei and Liang, Jian and Shi, Jie and Li, Houqiang and Ming, Gong and Duan, Nan},
  journal={arXiv preprint arXiv:2308.08089},
  year={2023}
}

@inproceedings{motionctrl,
  title={Motionctrl: A unified and flexible motion controller for video generation},
  author={Wang, Zhouxia and Yuan, Ziyang and Wang, Xintao and Li, Yaowei and Chen, Tianshui and Xia, Menghan and Luo, Ping and Shan, Ying},
  booktitle={ACM SIGGRAPH 2024 Conference Papers},
  pages={1--11},
  year={2024}
}

@inproceedings{draganything,
author = {Wu, Weijia and Li, Zhuang and Gu, Yuchao and Zhao, Rui and He, Yefei and Zhang, David Junhao and Shou, Mike Zheng and Li, Yan and Gao, Tingting and Zhang, Di},
title = {DragAnything: Motion Control for Anything Using Entity Representation},
year = {2024},
isbn = {978-3-031-72669-9},
publisher = {Springer-Verlag},
address = {Berlin, Heidelberg},
url = {https://doi.org/10.1007/978-3-031-72670-5_19},
doi = {10.1007/978-3-031-72670-5_19},
booktitle = {Computer Vision-ECCV 2024: 18th European Conference, Milan, Italy, September 29-October 4, 2024, Proceedings,  Part XXII},
pages = {331-348},
numpages = {18},
location = {Milan, Italy}
}

@article{tora,
  title={Tora: Trajectory-oriented Diffusion Transformer for Video Generation},
  author={Zhang, Zhenghao and Liao, Junchao and Li, Menghao and Qin, Long and Wang, Weizhi},
  journal={arXiv preprint arXiv:2407.21705},
  year={2024}
}

@inproceedings{motion-i2v,
  title={Motion-i2v: Consistent and controllable image-to-video generation with explicit motion modeling},
  author={Shi, Xiaoyu and Huang, Zhaoyang and Wang, Fu-Yun and Bian, Weikang and Li, Dasong and Zhang, Yi and Zhang, Manyuan and Cheung, Ka Chun and See, Simon and Qin, Hongwei and others},
  booktitle={ACM SIGGRAPH 2024 Conference Papers},
  pages={1--11},
  year={2024}
}

@ARTICLE{niqe,
  author={Mittal, Anish and Soundararajan, Rajiv and Bovik, Alan C.},
  journal={IEEE Signal Processing Letters}, 
  title={Making a “Completely Blind” Image Quality Analyzer}, 
  year={2013},
  volume={20},
  number={3},
  pages={209-212},
  doi={10.1109/LSP.2012.2227726}}

@INPROCEEDINGS{dynamics-dcc,
  author={Yin, Shanzhi and Zhang, Zihan and Chen, Bolin and Wang, Shiqi and Ye, Yan},
  booktitle={2025 Data Compression Conference (DCC)}, 
  title={Compressing Scene Dynamics: A Generative Approach}, 
  year={2025},
  volume={},
  number={},
  pages={414-414},
  doi={10.1109/DCC62719.2025.00101}}

@inproceedings{dcvc-rt,
  title={Towards practical real-time neural video compression},
  author={Jia, Zhaoyang and Li, Bin and Li, Jiahao and Xie, Wenxuan and Qi, Linfeng and Li, Houqiang and Lu, Yan},
  booktitle={Proceedings of the Computer Vision and Pattern Recognition Conference},
  pages={12543--12552},
  year={2025}
}

@INPROCEEDINGS{mr-dac,
  author={Konuko, Goluck and Valenzise, Giuseppe},
  booktitle={2024 IEEE 26th International Workshop on Multimedia Signal Processing (MMSP)}, 
  title={Multi-Reference Generative Face Video Compression with Contrastive Learning}, 
  year={2024},
  volume={},
  number={},
  pages={1-6},
  doi={10.1109/MMSP61759.2024.10743797}}

@INPROCEEDINGS{progressive-gfvc,
  author={Chen, Bolin and Yin, Shanzhi and Zhang, Zihan and Chen, Jie and Liao, Ru-Ling and Zhu, Lingyu and Wang, Shiqi and Ye, Yan},
  booktitle={2025 Data Compression Conference (DCC)}, 
  title={Beyond GFVC: A Progressive Face Video Compression Framework with Adaptive Visual Tokens}, 
  year={2025},
  volume={},
  number={},
  pages={163-172},
  doi={10.1109/DCC62719.2025.00024}}

@ARTICLE{mttf,
  author={Yin, Shanzhi and Chen, Bolin and Wang, Shiqi and Ye, Yan},
  journal={IEEE Transactions on Circuits and Systems for Video Technology}, 
  title={Generative Human Video Compression with Multi-granularity Temporal Trajectory Factorization}, 
  year={2025},
  volume={},
  number={},
  pages={1-1},
  doi={10.1109/TCSVT.2025.3596815}}

@INPROCEEDINGS{ihvc,
      title={Compressing Human Body Video with Interactive Semantics: A Generative Approach}, 
      author={Bolin Chen and Shanzhi Yin and Hanwei Zhu and Lingyu Zhu and Zihan Zhang and Jie Chen and Ru-Ling Liao and Shiqi Wang and Yan Ye},
      year={2025},
      booktitle={2022 IEEE International Conference on Image Processing},

}

@misc{imt,
      title={Rethinking Generative Human Video Coding with Implicit Motion Transformation}, 
      author={Bolin Chen and Ru-Ling Liao and Jie Chen and Yan Ye},
      year={2025},
      eprint={2506.10453},
      archivePrefix={arXiv},
      primaryClass={cs.CV},
      url={https://arxiv.org/abs/2506.10453}, 
}

@INPROCEEDINGS{imf,
  author={Gao, Yue and Li, Jiahao and Chu, Lei and Lu, Yan},
  booktitle={2024 IEEE/CVF Conference on Computer Vision and Pattern Recognition (CVPR)}, 
  title={Implicit Motion Function}, 
  year={2024},
  volume={},
  number={},
  pages={19278-19289},
  doi={10.1109/CVPR52733.2024.01824}}

@misc{gfvc-survey-trans,
      title={Generative Models at the Frontier of Compression: A Survey on Generative Face Video Coding}, 
      author={Bolin Chen and Shanzhi Yin and Goluck Konuko and Giuseppe Valenzise and Zihan Zhang and Shiqi Wang and Yan Ye},
      year={2025},
      eprint={2506.07369},
      archivePrefix={arXiv},
      primaryClass={cs.CV},
      url={https://arxiv.org/abs/2506.07369}, 
}

@ARTICLE{gfvc-sei-trans,
  author={Bolin Chen and Yan Ye and Jie Chen and Ru-Ling Liao and Shanzhi Yin and Shiqi Wang and Kaifa Yang and Yue Li and Yiling Xu and Ye-Kui Wang and Shiv Gehlot and Guan-Ming Su and Peng Yin and Sean McCarthy and Gary J. Sullivan},
  journal={IEEE Transactions on Multimedia}, 
  title={Standardizing Generative Face Video Compression using Supplemental Enhancement Information}, 
  year={2025}}

@misc{sparse2dense,
      title={Sparse2Dense: A Keypoint-driven Generative Framework for Human Video Compression and Vertex Prediction}, 
      author={Bolin Chen and Ru-Ling Liao and Yan Ye and Jie Chen and Shanzhi Yin and Xinrui Ju and Shiqi Wang and Yibo Fan},
      year={2025},
      eprint={2509.23169},
      archivePrefix={arXiv},
      primaryClass={cs.CV},
      url={https://arxiv.org/abs/2509.23169}, 
}

@misc{pleno,
      title={Pleno-Generation: A Scalable Generative Face Video Compression Framework with Bandwidth Intelligence}, 
      author={Bolin Chen and Hanwei Zhu and Shanzhi Yin and Lingyu Zhu and Jie Chen and Ru-Ling Liao and Shiqi Wang and Yan Ye},
      year={2025},
      eprint={2502.17085},
      archivePrefix={arXiv},
      primaryClass={cs.CV},
      url={https://arxiv.org/abs/2502.17085}, 
}

@inproceedings{llm_beates_diffusion,
title={Language Model Beats Diffusion - Tokenizer is key to visual generation},
author={Lijun Yu and Jose Lezama and Nitesh Bharadwaj Gundavarapu and Luca Versari and Kihyuk Sohn and David Minnen and Yong Cheng and Agrim Gupta and Xiuye Gu and Alexander G Hauptmann and Boqing Gong and Ming-Hsuan Yang and Irfan Essa and David A Ross and Lu Jiang},
booktitle={The Twelfth International Conference on Learning Representations},
year={2024},
url={https://openreview.net/forum?id=gzqrANCF4g}
}

@INPROCEEDINGS {controllale_fluid,
author = { Mahapatra, Aniruddha and Kulkarni, Kuldeep },
booktitle = { 2022 IEEE/CVF Conference on Computer Vision and Pattern Recognition (CVPR) },
title = {{ Controllable Animation of Fluid Elements in Still Images }},
year = {2022},
volume = {},
ISSN = {},
pages = {3657-3666},
doi = {10.1109/CVPR52688.2022.00365},
url = {https://doi.ieeecomputersociety.org/10.1109/CVPR52688.2022.00365},
publisher = {IEEE Computer Society},
address = {Los Alamitos, CA, USA},
month =Jun}

@INPROCEEDINGS {animate_fluid,
author = { Holynski, Aleksander and Curless, Brian and Seitz, Steven M. and Szeliski, Richard },
booktitle = { 2021 IEEE/CVF Conference on Computer Vision and Pattern Recognition (CVPR) },
title = {{ Animating Pictures with Eulerian Motion Fields }},
year = {2021},
volume = {},
ISSN = {},
pages = {5806-5815},
doi = {10.1109/CVPR46437.2021.00575},
url = {https://doi.ieeecomputersociety.org/10.1109/CVPR46437.2021.00575},
publisher = {IEEE Computer Society},
address = {Los Alamitos, CA, USA},
month =Jun}

@INPROCEEDINGS {lvdm,
author = { Blattmann, Andreas and Rombach, Robin and Ling, Huan and Dockhorn, Tim and Kim, Seung Wook and Fidler, Sanja and Kreis, Karsten },
booktitle = { 2023 IEEE/CVF Conference on Computer Vision and Pattern Recognition (CVPR) },
title = {{ Align Your Latents: High-Resolution Video Synthesis with Latent Diffusion Models }},
year = {2023},
volume = {},
ISSN = {},
pages = {22563-22575},
doi = {10.1109/CVPR52729.2023.02161},
url = {https://doi.ieeecomputersociety.org/10.1109/CVPR52729.2023.02161},
publisher = {IEEE Computer Society},
address = {Los Alamitos, CA, USA},
month =Jun}

@ARTICLE{ssim,
  author={Zhou Wang and Bovik, A.C. and Sheikh, H.R. and Simoncelli, E.P.},
  journal={IEEE Transactions on Image Processing}, 
  title={Image quality assessment: from error visibility to structural similarity}, 
  year={2004},
  volume={13},
  number={4},
  pages={600-612},
  doi={10.1109/TIP.2003.819861}}




\end{document}